\documentclass[journal]{IEEEtran}
\usepackage{cite}
\usepackage{amsmath,amssymb,amsfonts}
\usepackage{algorithmic}
\usepackage{graphicx}
\usepackage{textcomp}
\usepackage{hyperref}
\usepackage{graphicx}

\usepackage{booktabs,makecell,tabularx}
\newcolumntype{C}{>{\centering\arraybackslash}X} 
\usepackage{lipsum} % filler text

\ifCLASSINFOpdf
\else
\fi
\begin{document}
\title{High-Quality Face Caricature via Style Translation}

\author{Lamyanba Laishram,
        Muhammad Shaheryar,
        Jong Taek Lee,
        and~Soon Ki Jung,~\IEEEmembership{Senior Member,~IEEE}% <-this % stops a space
%\thanks{L. Laishram is with the School of Computer Science and Engineering, Kyungpook National University, Daegu, South Korea.}% <-this % stops a space
}

\maketitle

\begin{abstract}
Caricature is an exaggerated form of artistic portraiture that accentuates unique yet subtle characteristics of human faces. Recently, advancements in deep end-to-end techniques have yielded encouraging outcomes in capturing both style and elevated exaggerations in creating face caricatures. Most of these approaches tend to produce cartoon-like results that could be more practical for real-world applications. In this study, we proposed a high-quality, unpaired face caricature method that is appropriate for use in the real world and uses computer vision techniques and GAN models. We attain the exaggeration of facial features and the stylization of appearance through a two-step process: Face caricature generation and face caricature projection. The face caricature generation step creates new caricature face datasets from real images and trains a generative model using the real and newly created caricature datasets. The Face caricature projection employs an encoder trained with real and caricature faces with the pretrained generator to project real and caricature faces. We perform an incremental facial exaggeration from the real image to the caricature faces using the encoder and generator's latent space. Our projection preserves the facial identity, attributes, and expressions from the input image. Also, it accounts for facial occlusions, such as reading glasses or sunglasses, to enhance the robustness of our model. Furthermore, we conducted a comprehensive comparison of our approach with various state-of-the-art face caricature methods, highlighting our process's distinctiveness and exceptional realism.
\end{abstract}


\IEEEpeerreviewmaketitle

\section{introduction}

% What is a caricature and its applications:
A caricature is a visual portrayal of a person that simplifies or exaggerates their most visible characteristics through sketches or creative drawings \cite{sadimon2010computer}, which are primarily used to express humor and for entertainment. In traditional practice, caricatures are manually created by artists who carefully analyze the variations between an
individual’s unique features and the standard human facial characteristics. It is becoming more intriguing and essential to explore the automated generation of caricatures from given real images, as crafting a caricature demands significant effort, labor, and skill from the artist.

Computer vision applications encompass a broad spectrum, including the capability to create caricatures without requiring an artist’s direct involvement. Much like the process artists employ when creating caricatures, a computer vision-based approach can also be divided into two key phases. Firstly, it involves identifying distinctive features and enhancing them, and secondly, infusing the exaggerated image with artistic styles to match the artist’s preferences. This division into two independent categories adds flexibility and disentanglement, resulting in the creation of high-quality caricatures.

% How can it be produced or ways to create caricatures

The automated generation of a caricature from real images is a non-trivial challenge. Apart from imbuing the photo with a texture style reminiscent of caricatures, we should also take spatial exaggerations into consideration \cite{carigans, shi2019warpgan}. Previous methods for creating facial caricatures required the expertise of professionals to achieve satisfactory outcomes \cite{akleman2000making}. The issue of exaggerating facial features remains an open problem in research areas like detection \cite{yaniv2019face} and recognition \cite{shin2007combination}.

Certain methods incorporate additional data, such as user interaction \cite{liang2002example} or by increasing the shape representation's divergence from the average, as in the case of 2D landmarks or 3D meshes \cite{brennan1985caricature, mo2004improved, han2018caricatureshop, wu2018alive} to tackle the exaggeration challenge.
With the advancement in computer vision techniques, numerous automated caricature generation methods accomplish the exaggeration task by employing deep neural networks in an image-to-image translation manner \cite{han2018caricatureshop, shi2019warpgan, gong2020autotoon, cao2018carigans}. Certain approaches employ point-based warping techniques to convert real images into caricatures \cite{shi2019warpgan}.

Furthermore, there has been considerable research into automatic portrait style transfer, which is based on image style transfer \cite{li2016combining, selim2016painting, liao2017visual} and image-to-image translation \cite{kim2019u}. Deep learning techniques have been very successful in performing image translation by learning from representation data examples \cite{hinton2006reducing, huang2018multimodal}. Unfortunately, paired real and caricature are not commonly found. Training the translation process in a supervised manner is not practical, and the creation of such a dataset can be a laborious task. One of the readily accessible caricature datasets is WebCaricature \cite{huo2017webcaricature}, encompassing 6042 caricatures and 5974 photographs spanning 252 distinct identities. However, it's worth noting that the dataset's quality is subpar, with caricatures exhibiting inconsistent styles and exaggerations.

Due to the scarcity of paired image data, image-to-image translation is increasingly shifting towards training with unpaired images \cite{choi2018stargan, huang2018multimodal, zhu2017unpaired}, as well as gaining insights from unpaired portraits and caricatures \cite{cao2018carigans, wu2019attribute}.
Several studies  \cite{huang2018multimodal, liu2017unsupervised} have introduced unsupervised cross-domain image translation approaches, aiming to learn both geometric deformation and appearance translation simultaneously. 
However, training on unpaired images may introduce significant variations in exaggerations due to the substantial gap in shape and appearance between real and caricature images, often leading to unsatisfactory outcomes. Additionally, differences in poses and scales among images can make it challenging to differentiate facial features.

Neural style transfer techniques employ deep neural networks to transfer artistic styles from a reference to images and excel in stylizing appearances but do not enhance the geometric features \cite{johnson2016perceptual, liaovisual2017}. However, the advancement of Generative Adversarial Networks (GANs) \cite{goodfellow2020generative} has led to the emergence of state-of-the-art face generators like StyleGAN \cite{karras2019style, Karras_2020_CVPR}, which offer disentangled and high-fidelity images through transfer learning. 

\begin{figure*}[]    
    \centering
    \includegraphics[scale=0.32]{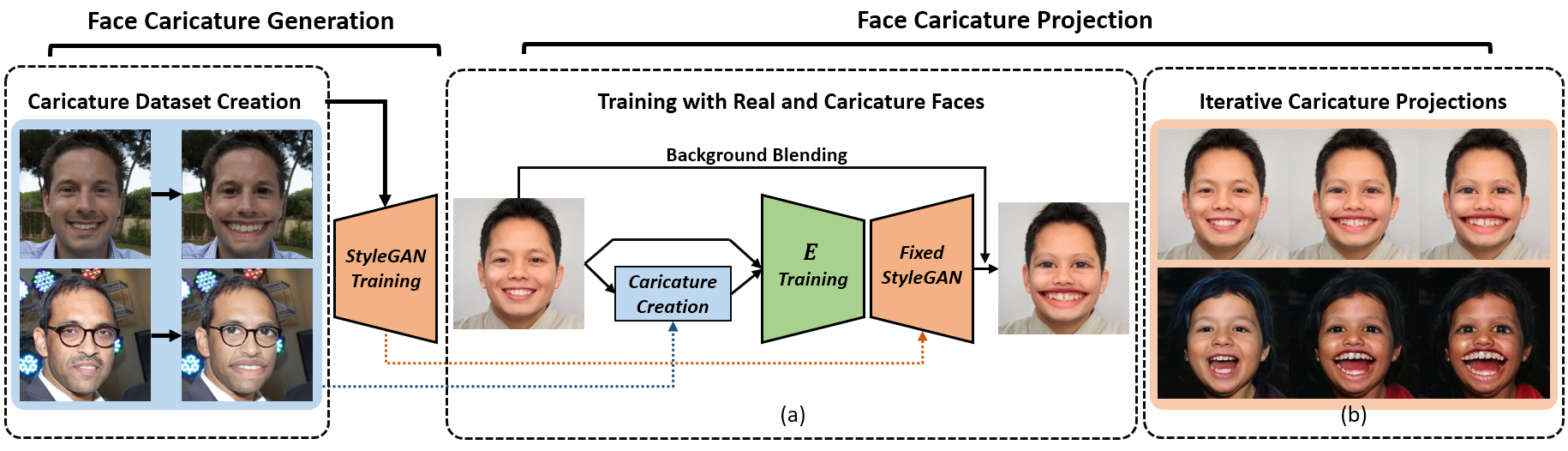}
    \caption{We present the overview of our proposed method. Our method consists of two key steps: Face Caricature Generation and Face Caricature Projection. In the first step, Face Caricature Generation, we create a caricature dataset from real faces. A generative model is trained with real and caricature faces, which can produce face caricatures and real images with different styles. The second step, Face Caricature Projection, involves training an encoder using the pretrained StyleGAN. (a) The encoder training process uses real and newly created caricature faces. (b) The first row shows the incremental facial exaggeration from real to caricature faces, and the second row shows the style change with facial exaggeration.}
    \label{fig: overview}
\end{figure*}

% How is our caricature different 
Our face caricature approach is different from the previous methods. Our main goal is to exaggerate facial features while keeping them realistic and usable in real-world scenarios.
Following the work in \cite{laishram2023style}, our caricature exaggerates the eye and mouth regions, keeping the face contour and other facial features unchanged. We create our realistic caricatures with the focus on three goals: (1) realistic face caricature with exaggerated eyes and mouth region, (2) making sure our caricature identity is the same as the input face,(3) our caricature should be realistic enough to be usable in real-world scenarios, and (4) unconditional visual style transfers and conserving all facial attribute from the real image to the caricature faces.

We proposed a novel caricature creation with a realistic style applicable to the real world. Style translation refers to the conversion of one style representation to another style representation. We utilize an unpaired caricature learning method to achieve our goal. We exaggerate facial features and the stylization of appearance through a two-step process: face caricature generation and face caricature projection. The initial phase of the face caricature generation step is the creation of new caricature face datasets from real images. We then train a style generator using the real and the new caricature datasets, discussed in Section \ref{lb: Caricature dataset creation}. The face caricature projection employs an encoder which is trained with our pretrained generator. The encoder is trained using real and our new caricature images to project similar real and caricature faces. Additionally, using the projected real and caricature images, we achieve an incremental facial exaggeration from the real to the caricature images,  which provides flexibility in our method.
The projection of the real and caricature images preserves the facial identity, attributes, and expressions from the input images, discussed in Section \ref{lb: face caricature projection}. Our method also addresses facial occlusions like reading glasses or sunglasses to enhance the robustness of our model.

This work presents several significant contributions:
\begin{enumerate}
\item
Our approach employs an unpaired learning procedure to produce caricatured faces from real facial images. We don't require a pair of real and caricature images. It accomplishes both facial exaggeration and style transfer from the real face image.
\item We produce caricature face datasets from real face images. We train StyleGAN using the real and the new caricature faces, enabling the synthesis of different styles of real and caricature faces. We further designed an encoder to get the full translation of expressions, poses, and attributes from the real faces.
\item
Our caricature projection provides an incremental exaggeration of facial features. This incremental process provides flexibility in our caricature projection as the extent of facial exaggeration can be performed according to one's preference.
\item
Our generated caricatures exhibit superior realism and quality compared to state-of-the-art methods. Our caricatures maintain high quality, making them more visually convincing when used in real-world scenarios. 
\end{enumerate}

\vspace{2mm}

\noindent The remainder of the paper is structured as follows: Section \ref{lb: related work} provides the necessary face caricature background and reviews recent advances in facial image generation and StyleGAN inversion; Section \ref{lb: caricature creation} outlines the two-stage proposed framework, explaining face caricature generation and face caricature projection; Section \ref{lb: implementation} details implementation settings and dataset uses; Section \ref{lb: experiment} details experiment settings and evaluates the results; and Section \ref{lb: conclusion} concludes this paper with discussions.

%%%%%%%%%%%%%%%%%%%%% RELATED WORK %%%%%%%%%%%%%%%%%%%%%%%

\section{Related Work}
\label{lb: related work}

\subsection{Caricature Creation}

Creating caricatures entails the recognition and exaggeration of unique facial characteristics while preserving the individual's identity. Caricatures can be crafted using three approaches: distorting facial attributes, employing style transfer, or utilizing methods that combine both techniques.

Conventional techniques operate by amplifying the deviation from the average, achieved through methods such as explicitly identifying and warping landmarks \cite{gooch2004human,liao2004automatic} or employing data-driven approaches to estimate unique facial attributes \cite{liu2006mapping, zhang2016data}. As generative networks have advanced, some image-to-image translation methods \cite{zheng2019unpaired,li2020carigan} have been undertaken to incorporate style transfer. Nevertheless, these networks are unsuitable for applications involving significant spatial variations, resulting in outputs with diminished visual quality. Zhang et al. \cite{zhang2021disentangled} introduced an approach that aims to acquire a disentangled feature representation of various facial attributes, enabling the generation of realistic portraits that exhibit appropriate exaggerations and a rich diversity in style.

Cao et al. \cite{cao2018carigans} employ two CycleGANs, trained on image and landmark spaces, to handle texture rendering and geometry deformation. WarpGAN \cite{shi2019warpgan} surpasses visual quality and shape exaggeration, providing flexibility in spatial variability for both image geometry and texture. CariGAN \cite{cao2018carigans}, on the other hand, is a GAN trained with unpaired images, focusing on learning image-to-caricature translation. Shi et al. \cite{shi2019warpgan} introduce an end-to-end GAN framework that simultaneously trains warping and style. AutoToon \cite{gong2020autotoon} utilizes deformation fields to apply exaggerations and is trained in a supervised manner using paired data derived from artist-warped photos to learn warping fields. However, it is limited to mapping to a single domain, making it unable to produce diverse exaggerations. Abdal et al. \cite{abdal20233davatargan} introduced a technique for crafting 3D caricatures that permits the modification and animation of personalized artistic 3D avatars using artistic datasets.

\subsection{Style Transfer}

One aspect of image synthesis that poses a challenge is style transfer, which aims to create a content image with multiple styles. Thanks to the practical ability of convolutional neural networks (CNNs) \cite{gatys2017controlling} to extract semantic features, numerous networks dedicated to style transfer have been developed. The initial style rendering process was introduced by Gatys et al. \cite{gatys2015texture}, who employed hierarchical features from a VGG network \cite{simonyan2014very}. Gatys et al. \cite{gatys2016image} pioneered the first neural style transfer approach, utilizing a CNN to transfer style information from a style image to a content image. However, a drawback of this approach is that the style and content need to be similar, which is different from caricature images.

A promising area of research lies in the application of Generative Adversarial Networks (GANs) \cite{goodfellow2020generative} for image synthesis, which has yielded cutting-edge results in various domains such as text-to-image translation \cite{reed2016generative} and image inpainting \cite{yeh2016semantic}.
Regarding unpaired image translation, methods like CycleGAN \cite{zhu2017unpaired} have been utilized, leveraging a cycle consistency loss to achieve translation between different image domains. Additionally, approaches like StarGAN  \cite{choi2018stargan,choi2020stargan} employ a single generator to learn mappings across various image domains.
However, capturing the geometric transformations required for direct photo-to-caricature mapping in an image-to-image translation framework remains a challenging task.

StyleGAN \cite{karras2019style},\cite{karras2020training} excels at producing high-fidelity facial images with fine-grained control over hierarchical channel styles. Many techniques leverage StyleGAN for the generation of high-quality images and the manipulation of facial characteristics, as well as for various applications related to faces, including swapping, restoration, de-aging, and reenactment \cite{melnik2022face, xia2022gan}.
Pinkney and Adler \cite{pinkney2020resolution} further enhanced StyleGAN using sparse cartoon data, demonstrating its effectiveness in generating lifelike cartoon faces. Additionally, DualStyleGAN \cite{yang2022pastiche} provides customizable control over dual style transfers, catering to both the extended artistic portrait domain and the original face domain. StyleCariGAN \cite{jang2021stylecarigan} produced shape exaggeration and stylization by mixing layers of photo and caricature styles.

%%%%%%%%%%%%%%%%%%%%% Style based caricature creation %%%%%%%%%%%%%%%%%%%%%%%

\begin{figure*}[]
  \centering
  \includegraphics[scale=0.32]{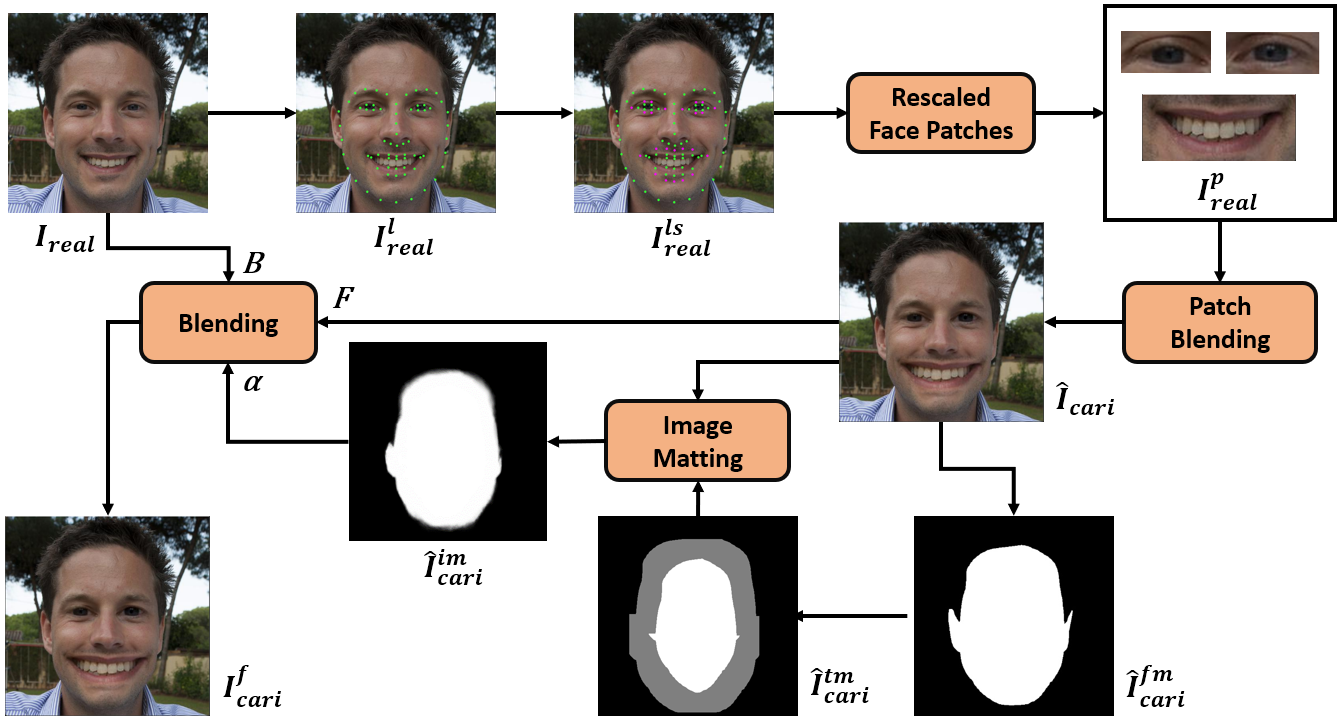}
  \caption{The steps for our face caricature data creation. We perform multiple image operations techniques to achieve our face caricature dataset.}
  \label{fig: caricature creation}
\end{figure*}

\section{Style based caricature creation}
\label{lb: caricature creation}

Our proposed method operates transparently and understandably, comprising two distinct stages: face caricature generation and face caricature projection, as illustrated in Figure \ref{fig: overview}.
In the initial stage, our focus is creating new facial caricature datasets, which exaggerate eyes and mouth regions while preserving the facial contours. Subsequently, we train a style generator called StyleGAN \cite{karras2019style} using the real and our new caricature datasets, which can generate highly realistic images in different styles. In the second stage, we design a projection model to produce high-quality caricature faces from real facial images and the incremental exaggeration of facial features. Our proposed method projects the genuine facial image into a caricature representation, emphasizing the unique and exaggerated facial features that constitute an individual's appearance.

\subsection{Background}

\subsubsection{StyleGAN} We use StyleGAN2’s \cite{karras2020training}, a style-based network that controls the synthesis of images. To train StyleGAN, a large dataset of real images is used, which is then processed to learn the underlying patterns and characteristics of the data. We use real and caricature faces for our method. The model learns to generate real and caricature images visually similar to those in the training set.

\subsubsection{Projection Techniques}
We can project an input image into an equivalent output image using the StyleGAN architecture by employing two distinct approaches: latent code optimization and encoder-based methods.
Our approach is predominantly focused on encoder-based methods for several compelling reasons. Firstly, these methods offer significant speed advantages, as they can map the latent code in a single forward pass. The encoder-based approach contrasts with the optimization-based approach, which can be computationally demanding for each image. Secondly, the output of an encoder resides within a compact and well-defined space, rendering it more suitable for subsequent editing and manipulation tasks.

\subsubsection{Nature of StyleGAN Latent space}
The StyleGAN latent space plays a crucial role in creating and manipulating the characteristics of the generated real and caricature faces. The latent space vectors control various aspects of image generation, like facial features, colors, textures, etc. Another characteristic of StyleGAN latent space is the nature of disentanglement, where each latent space direction corresponds to specific features or attributes of the generated image. A smooth interpolation between two points in the latent space creates images that transition between different attributes.

\subsection{Face Caricature Generation}
\label{lb: Caricature dataset creation}
In the face caricature generation, datasets are formed by creating exaggerated facial representations featuring enlarged eyes and mouths using real face images. Here, we also discuss generating exaggerated faces with facial obstacles, like the faces with eyeglasses. After creating caricature faces, We followed the generation process by training a style generator with our caricature face dataset.

\begin{figure}[h]
  \centering
  \includegraphics[scale=0.12]{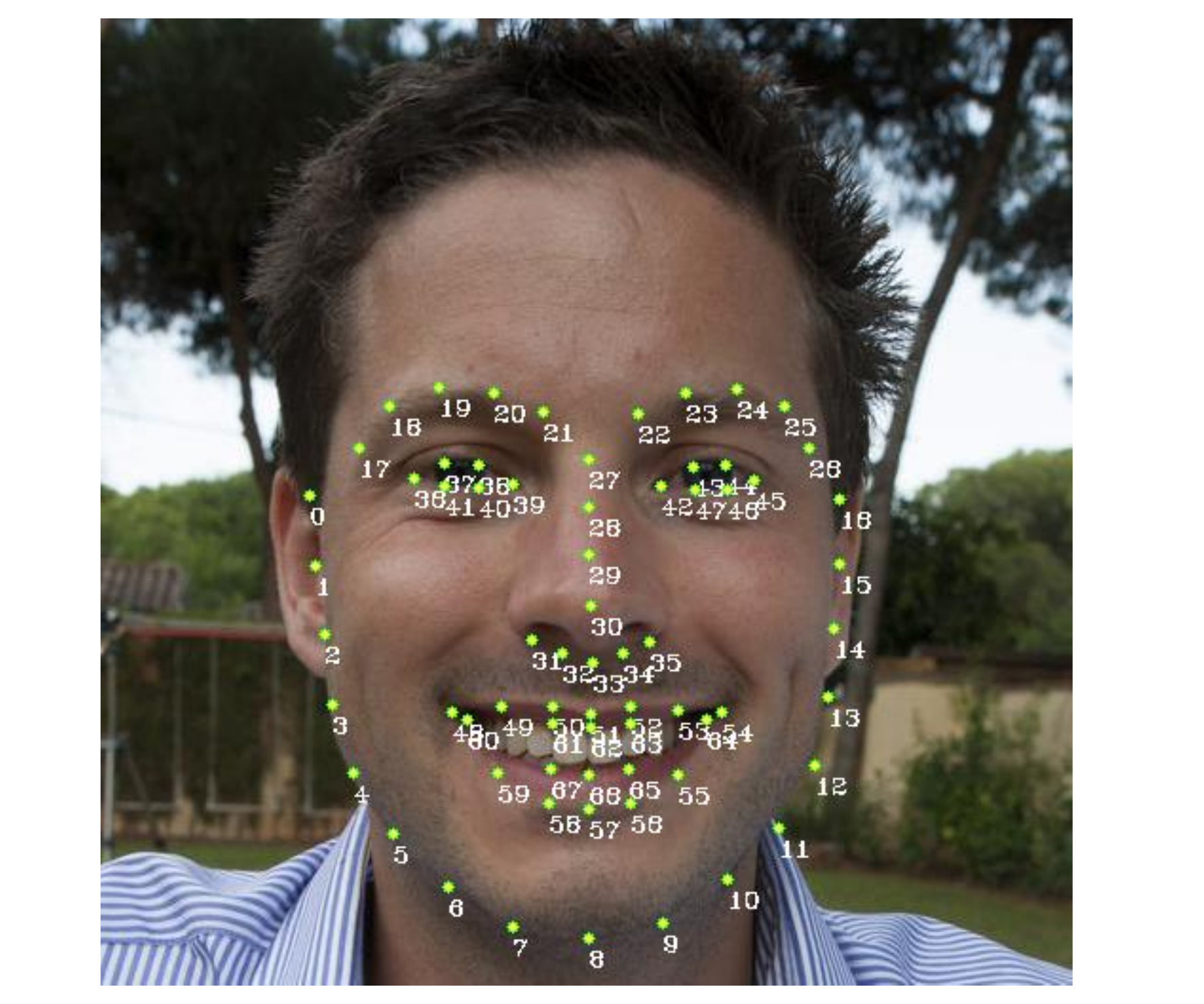}
  \caption{Face landmarks representation: The landmark position with the 68 indexes representation on the FFHQ dataset.}
  \label{fig: landmark}
\end{figure}

\subsubsection{Face Caricature Dataset Creation}
\label{lb: caricature dataset}

To create our caricature faces, we utilize real-face images randomly sampled from the FFHQ \cite{karras2019style} and CelebA-HQ \cite{liu2015deep} dataset. Both datasets provide a diverse range of genders, races, ages, expressions, and poses, ensuring the variety and representation of our caricature faces. The pipeline for caricature creation is divided into three stages: (i) facial landmark enlargement, (ii) face patch rescaling, and (iii) image matting, as illustrated in Figure \ref{fig: caricature creation}.

In the first stage, we use landmark detectors to detect the facial landmarks in the real input image $I_{real}$. Specifically, we employ a pre-trained detector from the Dlib library \cite{Dlib}, which estimates the location map of the facial structure. The Dlib library detects 68 facial landmarks, each assigned specific (x, y) coordinates ranging from 0 to 67. These landmarks correspond to different parts of the face, such as the eyes, eyebrows, nose, mouth, and face contour, as depicted in Figure \ref{fig: landmark}. We represent the input face image with identified landmarks as $I^{l}_{real}$, and the coordinates are highlighted with green markers. This initial stage of landmark detection provides crucial information about the facial structure, enabling us to proceed to the subsequent steps of face patches and blending, as well as image matting and blending.

\begin{figure}[h]
  \centering
  \includegraphics[width=\linewidth]{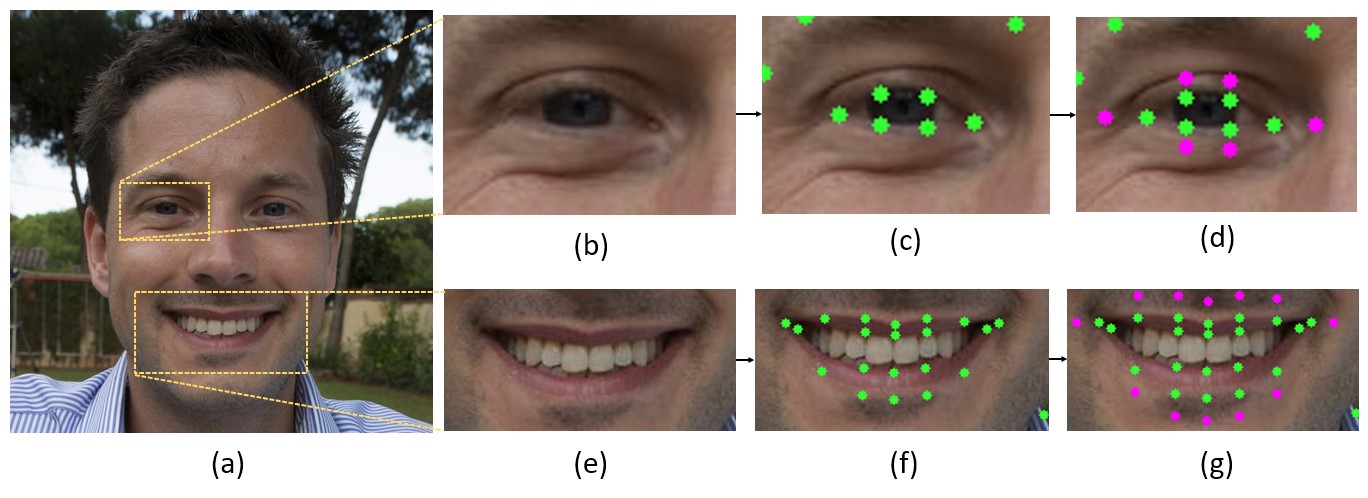}
  \caption{Face landmark enlargement for patch generation. (b-d) represents the eye landmark enlargement. (e-g) represents the mouth landmark enlargement. The landmarks produced by the detector are represented in green, and the enlarged landmarks are represented in pink.}
  \label{fig: patch}
\end{figure}

In the second stage, face patch rescaling, we perform several operations, including the production of face patches, the exaggeration of these patches, and blending them into the original image to create the caricature effect. Our focus for exaggeration is on the eyes and mouth regions of the face. To accurately target the eye regions, we group the landmark indexes into the left and right eyes, as depicted in Figure \ref{fig: landmark}. The mouth area consists of the upper and lower lips, and we consider specific landmark indexes for these regions. In the case of the mouth, we utilize the top landmark indexes for the upper lip and the bottom indexes for the lower lip. Using these landmark indexes, we produce face patches that will undergo exaggeration. We achieve this by enlarging the coordinates of the landmarks corresponding to the eye and mouth regions, as illustrated in Figure \ref{fig: patch}.
The resulting image $I^{ls}_{real}$ displays the face with enlarged landmark coordinates, which are highlighted in pink. To further enhance the exaggeration effect, we scale the face patches to a factor of 1.5, resulting in exaggerated face patches represented as $I^{p}_{real}$. These exaggerated patches seamlessly blend into the original image $I^{real}$.

\begin{figure}[]
  \centering
  \includegraphics[scale=0.25]{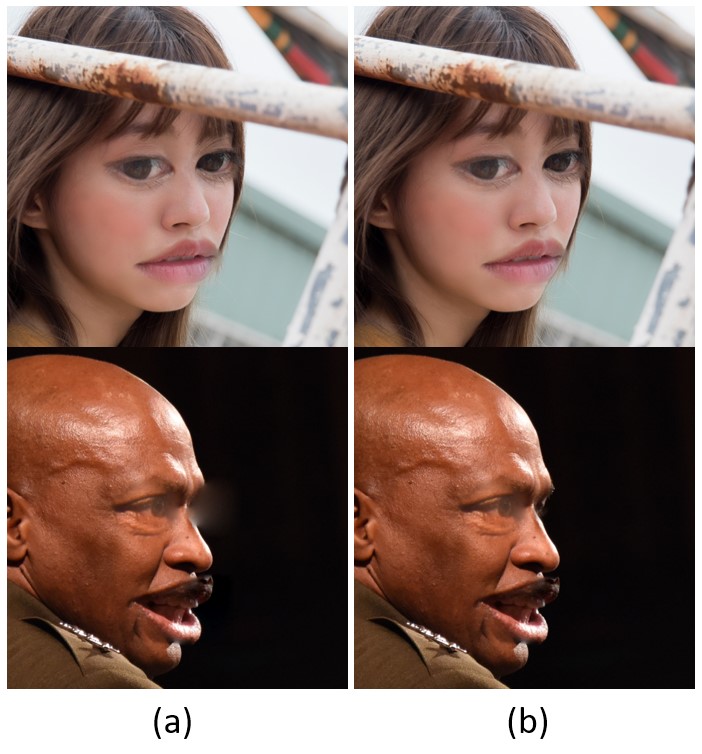}
  \caption{(a) The blurring of face patches after the blending process due to extreme head pose. (b) The result of removal of blurring after the image matting.}
  \label{fig: blur iluustration}
\end{figure}

\begin{figure*}[t]
  \centering
  \includegraphics[scale=0.35]{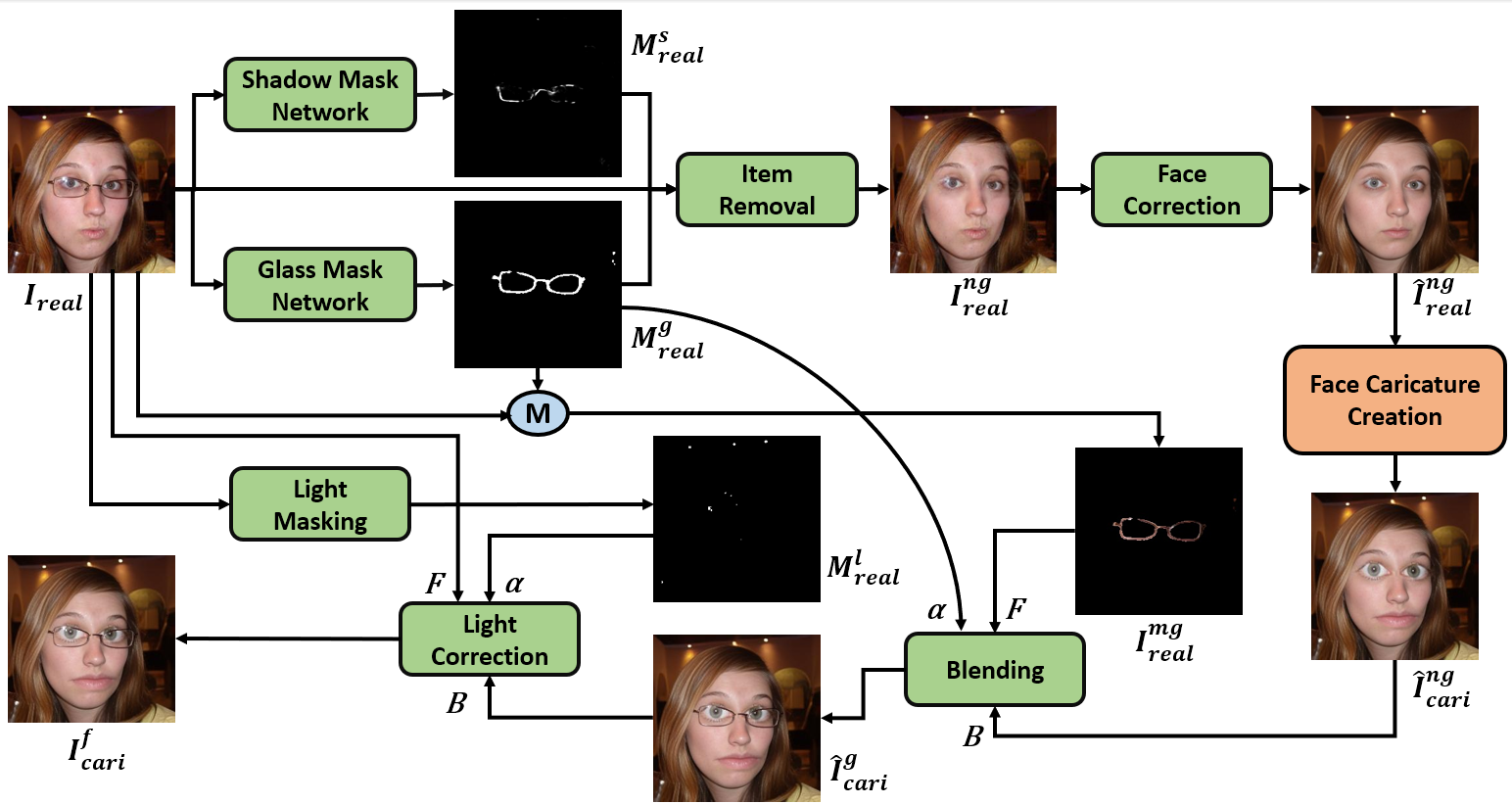}
  \caption{The workflow of creating the reading glass caricature face dataset. We perform additional steps to overcome the face occlusion challenges for reading glass caricature creation.}
  \label{fig: eyeglass}
\end{figure*} 

For the blending process, we employ the Poisson image editing technique \cite{perez2003poisson}, which ensures seamless and natural integration of the exaggerated patches with the original image. This technique considers factors such as image illumination and texture, resulting in a visually pleasing caricature effect.
By applying these operations, we can generate the final caricature image $\hat{I}_{cari}$, where the distinctive exaggerated features, such as enlarged eyes and mouth, seamlessly blend into the original face image while maintaining a natural appearance.
The Poisson editing method influences both image illumination and texture and is represented as follows:

\begin{equation}
\begin{split}
v=argmin_{v}\sum_{i\epsilon S, j\epsilon N_i \cap S} ( (v_{i}-v_{j}) - (s_{i}-s_{j}) )^{2} \\
+ \sum_{i\epsilon S, j\epsilon N_i \cap \neg S} ( (v_{i}-t_{j}) - (s_{i}-t_{j}) )^{2},
\label{eq: poisson}
\end{split}
\end{equation}

\noindent where $\upsilon$ represents the pixel values of the new image, $s$ corresponds to the pixel values of the source image, $t$ represents the pixel values of the target image, $S$ signifies the destination domain, and $N_i$ denotes a set of neighboring pixels of $i$.

In the third stage, we tackle the problem of blurriness that can occur along the contours of the face, especially in cases where faces exhibit extreme poses during the blending process in stage two. To mitigate this blurring effect, we apply an image matting technique. First, we generate face masks from the previously obtained caricature image $\hat{I}_{cari}$ using a face segmentation method \cite{yu2018bisenet}, resulting in a mask image $\hat{I}^{fm}_{cari}$. In this mask, the foreground corresponds to the face region, while the background encompasses the remaining areas. Notably, we perform segmentation only for the face region, excluding the hair, as blurring tends to occur mainly in the hair-background region.

Next, we generate a trimap mask $\hat{I}^{tm}_{cari}$ from the face mask $\hat{I}^{fm}_{cari}$, using trimap mask generation process \cite{gupta2016automatic}. It involves applying a series of erosion and expansion operations to the foreground region of the face mask, using specific parameter values tailored to our method. With the face caricature image $\hat{I}_{cari}$ and the trimap mask $\hat{I}^{tm}_{cari}$ in hand, we apply an image matting method \cite{park2022matteformer}. This technique effectively addresses the blurring issue by enhancing the sharpness and clarity of the face contours in the caricature image. The image matting process utilizes both the caricature image and the trimap mask to generate a refined caricature image $\hat{I}^{im}_{cari}$. By employing this image matting stage, we can improve the overall visual quality of the caricature image by reducing blurring effects around the face contours, resulting in a more polished and realistic appearance.

The blurring on the face contour is removed by performing alpha blending. The image alpha blending technique requires a foreground, a background, and an alpha mask. We set the $\hat{I}_{cari}$ as foreground, $I_{real}$ as background, and $\hat{I}^{im}_{cari}$ as alpha mask. The alpha blending can be performed using the following equation:

\begin{equation}
I_{p} =\alpha_{p}F_{p}+(1-\alpha_{p})B_{p},
\label{eq: alpha}
\end{equation}

\noindent where $\alpha_{p}$ denotes the matte and within the range value of [0,1], and $F_{p}$ and $B_{p}$ correspond to the pixel values for the foreground and background, respectively. When $\alpha_{p}$ = 1 or 0, it signifies that the pixel at that position unequivocally belongs to the foreground or background, respectively. Otherwise, such a pixel is termed a partial or mixed pixel. Following the ultimate blending procedure, we produce our caricatured face denoted as $\hat{I}^{f}_{cari}$. Figure \ref{fig: blur iluustration}  illustrates eliminating blurring after the matting process.

\subsubsection{Face Caricature Dataset with Occulsion}

For generating our caricature dataset, we use various images, including faces with eyeglass occlusions. We address the caricature generation for face occlusion caused by eyeglasses to enrich our caricature dataset. The faces with eyeglasses can be categorized into two: (i) Reading glasses and (ii) sunglasses. We organized all the transparent glasses as reading glasses and the remaining as sunglasses. 

\vspace{2mm}

\textbf{\textit{Face Caricature with Reading Glasses:}} The whole pipeline for reading glass caricature generation is shown in Figure \ref{fig: eyeglass}. We can divide the reading glass caricature generation into five stages: (i) glass removal, (ii) correction, (iii) caricature generation, (iv) putting back glasses, and (iv) lighting correction.

The first stage is glass removal, where we remove both the reading glass and the cast shadow from the face image. We employ two networks, Shadow Mask Network and Glass Mask Network \cite{lyu2022portrait}, for the glass removal. Given an input image $I_{real}$, we generate two masks: a glass mask $M^{g}_{real}$ using the glass mask network and a shadow mask $M^{s}_{real}$ using the shadow mask network. We use item removal from \cite{lyu2022portrait} to remove both the eyeglass and the shadow from the face image and generate a new face image $I^{ng}_{real}$ with no eyeglass and cast shadow.

The second stage is the correction stage, where information on the image that was lost during the item removal stage is retrieved. We use an image restoration method \cite{zhou2022towards} to restore the degraded image and restore lost details. The corrected image $\hat{I}^{ng}_{real}$ restores both quality and fidelity and shows robustness to the degraded parts.

The third stage is the caricature generation process. The face image now has no reading glasses, so it performs the caricature generation method discussed in the previous section and generates appropriate caricature $\hat{I}^{ng}_{cari}$.

We put the glasses from $I_{real}$ into $\hat{I}^{ng}_{cari}$ in the fourth stage. We first generate a glass image $I^{mg}_{real}$ with only glasses using a bitwise AND mask operation using $I_{real}$ and $M^{g}_{real}$. We perform the alpha blending presented in Equation \ref{eq: alpha} to put back the reading glass from $I_{real}$ in our caricature face. We set the $I^{mg}_{real}$ as foreground, $\hat{I}^{ng}_{cari}$ as background, and $M^{g}_{real}$ as alpha mask. The generated caricature face with reading glasses is represented as $\hat{I}^{g}_{cari}$.

The final step is the lighting correction. We must ensure the light illumination is preserved from $I_{real}$ during this whole caricature process. We generate a light mask $M^{l}_{real}$ from $I_{real}$ by keeping a specific threshold that only the illuminated area is highlighted. We perform the alpha blending technique in Equation \ref{eq: alpha} to retrieve the lost light illumination from $I_{real}$. We set $I_{real}$ as foreground, $\hat{I}^{g}_{cari}$ as background, and $M^{l}_{real}$ as alpha mask.
Finally, we create our reading glass caricature image, ${I}^{f}_{cari}$.

\begin{figure}[]
  \centering
  \includegraphics[width=\linewidth]{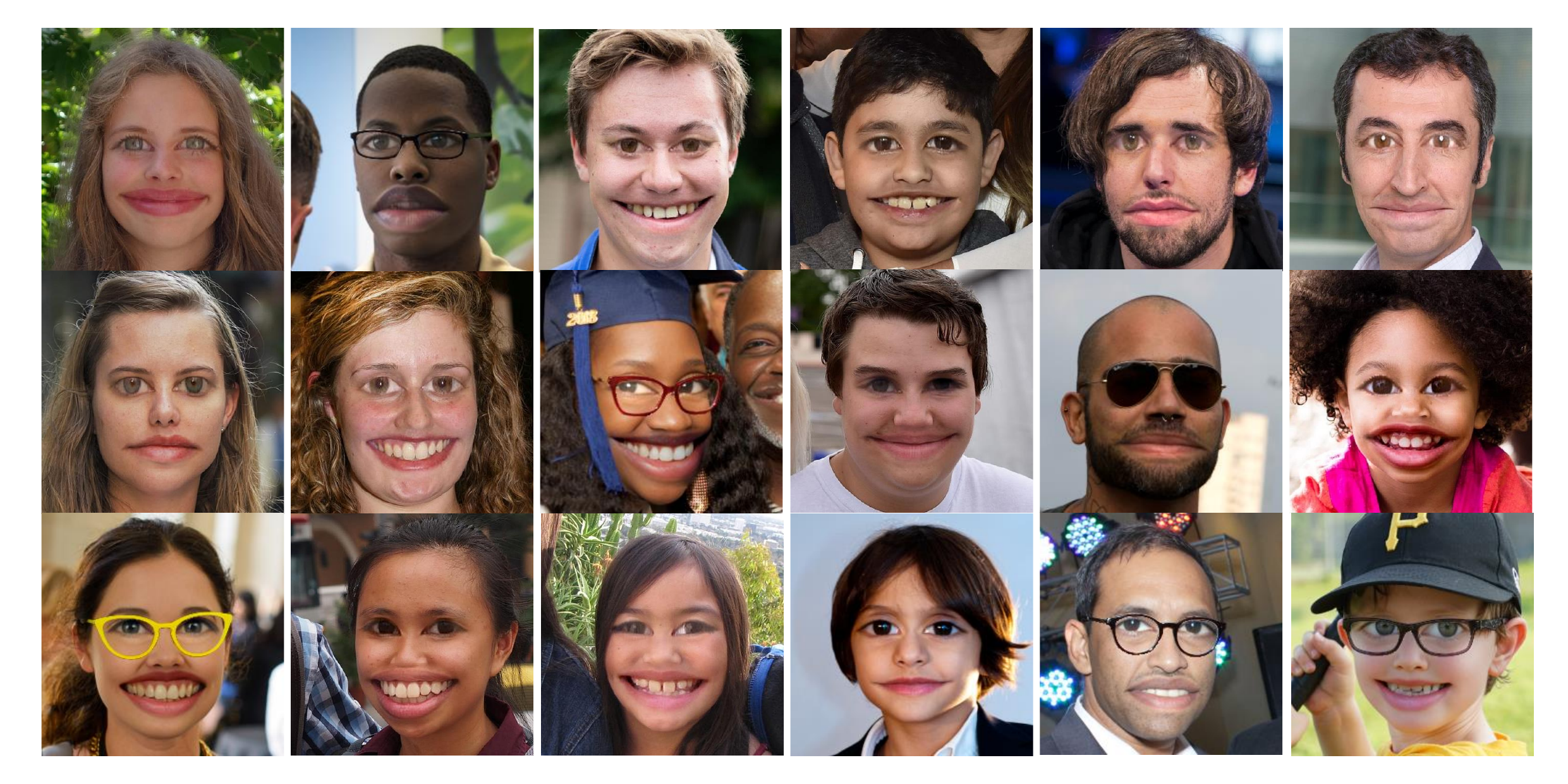}
  \caption{Examples of our dataset produced in the face caricature creation.}
  \label{fig: cari dataset}
\end{figure}

\textbf{\textit{Face Caricature with Sunglasses:}} We consider the face with sunglasses where it can't be see-through. The caricature generation of faces with sunglasses is a simple, straightforward process where we exaggerate face patches only for the mouth region. After the landmark detection in Figure \ref{fig: caricature creation}, only the mouth landmark has been enlarged, represented in Figure \ref{fig: patch}. The patch blending presented in Equation \ref{eq: poisson} is performed only for the mouth patch and generates a caricature face with sunglasses. The remaining steps are same as in Section \ref{lb: Caricature dataset creation}.

\begin{figure}[h]
  \centering
  \includegraphics[width=\linewidth]{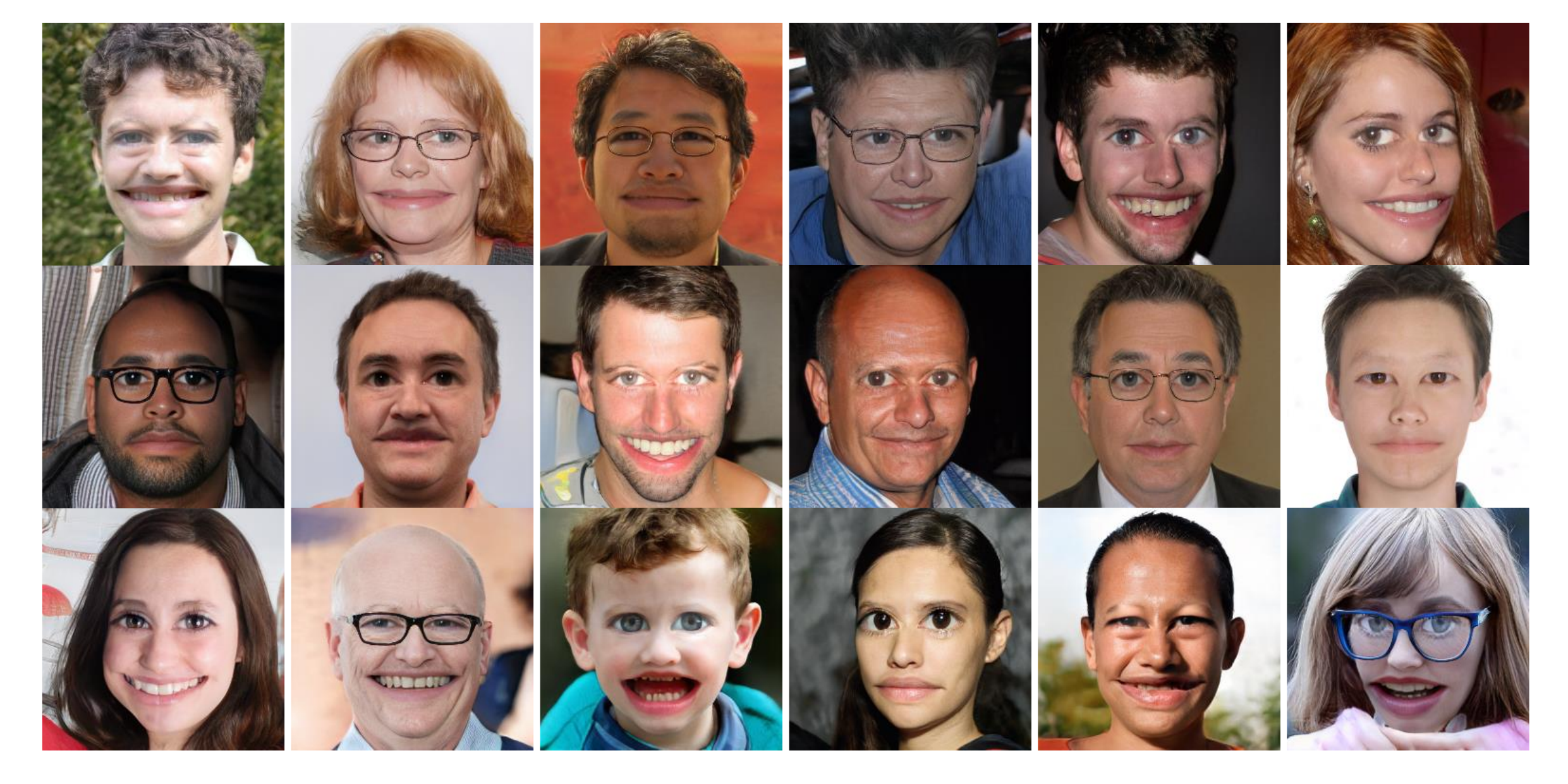}
  \caption{Random sample of caricature faces generated by the StyleGAN.}
  \label{fig: SGdata}
\end{figure}

\begin{figure}[h]
  \centering
  \includegraphics[width=\linewidth]{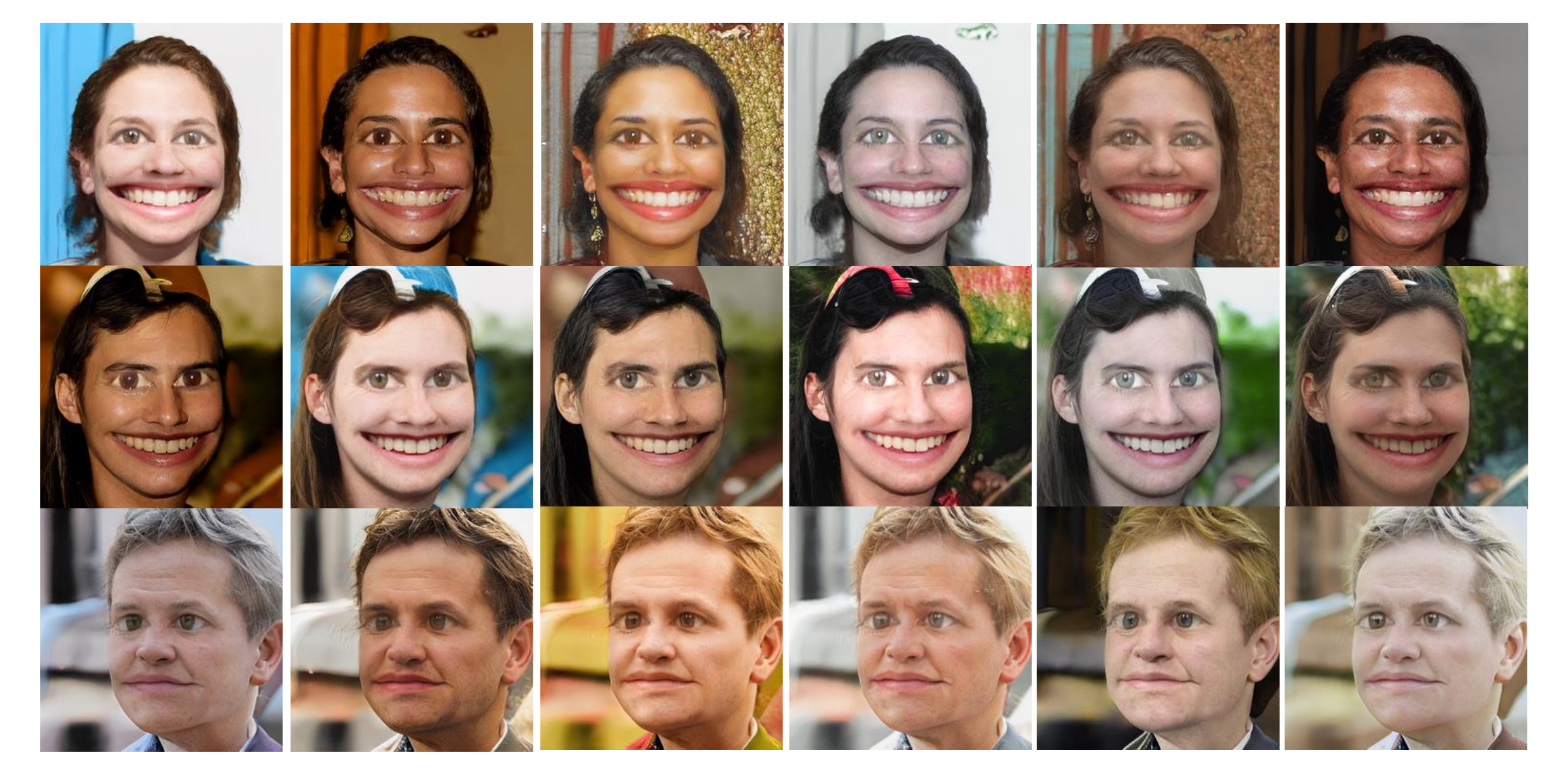}
  \caption{After StyleGAN training, we represent different styles for a specific face. Each row represents one identity.}
  \label{fig: SG style}
\end{figure}

\begin{figure*}[]
  \centering
  \includegraphics[scale=0.34]{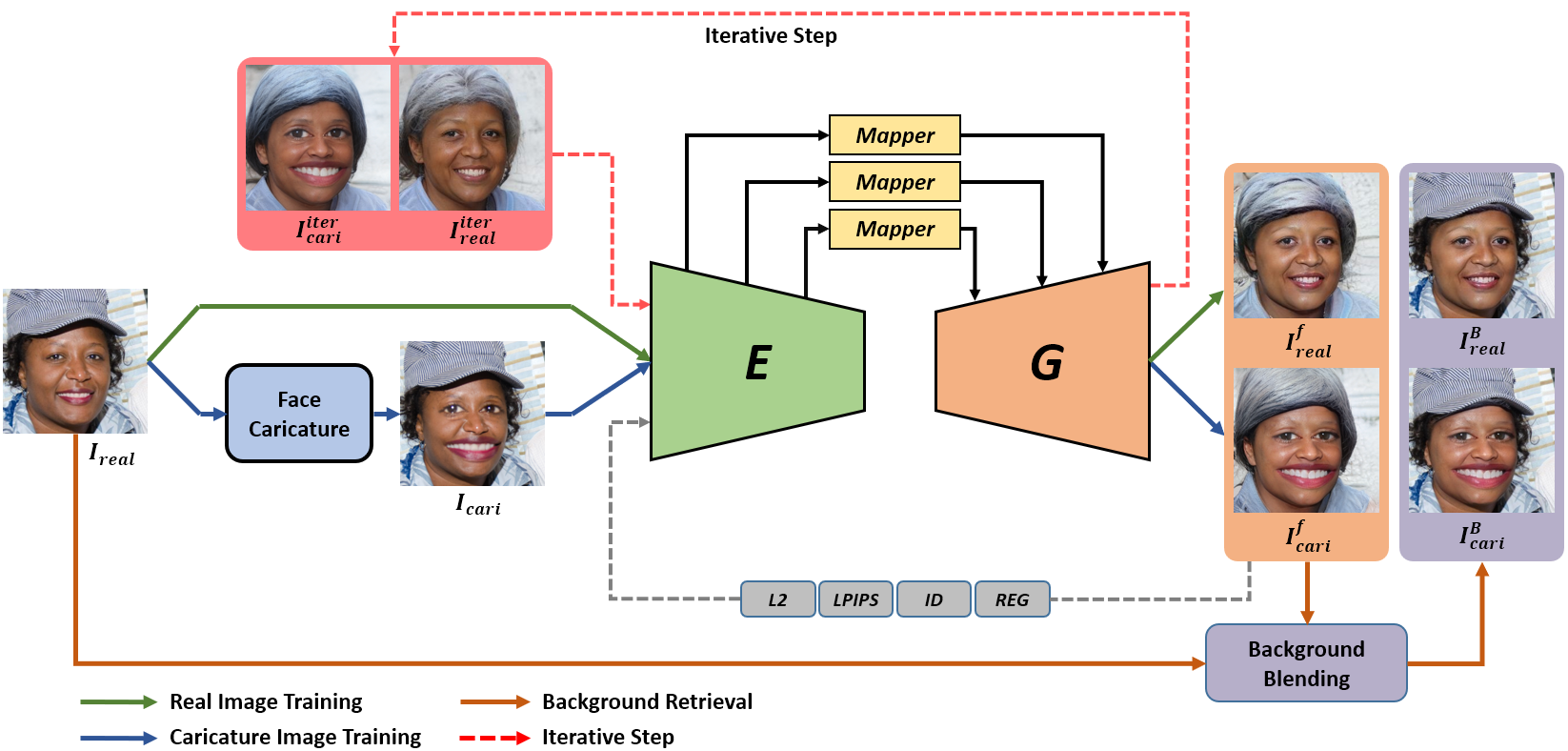}
  \caption{The encoder training for our Face caricature projection. We employ an encoder-based method with our pretrained StyleGAN, $G$. The encoder $E$ is trained using real and new caricature faces. We perform iterative steps to enhance the quality of our generated images and make them more faithful to the input faces. We perform a background blending process to get the background information the projected image cannot generate from the input image.}
  \label{fig: projector}
\end{figure*}

\subsubsection{Face Caricature Dataset}
We have successfully generated a diverse collection of caricature face images encompassing various attributes such as gender, race, age, expression, pose, illumination, etc. We use the FFHQ \cite{karras2019style} and CelebA-HQ \cite{liu2015deep} datasets for our caricature creation. Some examples of our caricature dataset are illustrated in Figure \ref{fig: cari dataset}.

\subsubsection{Style Generator}
\label{lb: stylegan}

The final step for the caricature generation is the training of StyleGAN \cite{karras2019style, Karras_2020_CVPR} architecture. The StyleGAN architecture comprises two networks: a mapping network and a synthesis network.
The mapping network, denoted as $f$, is an 8-layer Multi-Layer Perceptron (MLP) responsible for mapping a given latent code $z$ from the set $Z$ to generate $w$ in the set $W$. It can be represented as $ f: Z\to W $.
The synthesis network, $g$, consists of 18 convolutional layers, with each layer being controlled via adaptive instance normalization (AdaIN) \cite{huang2017arbitrary}. AdaIN incorporates the learned affine transformation ``A'' derived from the latent code $w$ at each layer. Additionally, a scalable Gaussian noise input ``B'' is introduced into each layer of the synthesis network $g$.

The architectural design ensures that each style influences only a single convolution. Random latent codes serve as a means to control the styles of the generated images. The StyleGAN training process exclusively used the real and newly created face caricature images. Following the training of StyleGAN, the generator can produce real and caricature images with diverse facial attributes, including variations in skin tone, hair color, shapes, and more.
It's crucial to underscore that our caricature generation generator stands out from previous approaches in a notable manner in terms of realism and usability. 
After training the StyleGAN, we generate random samples from the latent space to visualize how our caricature performs. The results are high quality and realistic, as shown in Figure \ref{fig: SGdata}. 
We can also generate different styles for different identities, and some examples are shown in Figure \ref{fig: SG style}.

\subsection{Face Caricature Projection}
\label{lb: face caricature projection}

Our caricature projection technique employs an encoder trained with two different datasets. The encoder is trained using real and caricature images with our pretrained StyleGAN from Section \ref{lb: stylegan}. The training framework of the encoder is shown in Figure \ref{fig: projector}.
For training the encoder, denoted as $E$, with our pretrained StyleGAN generator, represented as $G$, given an input source image $I_{real}$, we first create a corresponding caricature from the real input face, $I_{cari}$, following the process in Section \ref{lb: caricature dataset}. The newly created caricature faces with the real faces are used in the training of $E$ with the primary objective of $I^{f}_{cari}$ = $G(E(I_{cari}))$, such that $I^{f}_{cari} \approx  I_{cari}$ and $I^{f}_{real}$ = $G(E(I_{real}))$, such that $I^{f}_{real} \approx  I_{real}$. To enhance the quality of our generated images and make them more faithful to the input, we perform two forward passes through the encoder, $E$, and generator $G$.
Our goal is to efficiently and effectively produce high-quality real and caricatured faces, all while preserving the desired characteristics and visual resemblance to the input images.

Follows a methodology similar to the PSP \cite{richardson2021encoding} and e4e \cite{tov2021designing} approaches. We utilize a Feature Pyramid Network \cite{lin2017feature} built upon a ResNet \cite{he2016deep} backbone, extracting style features from three intermediate levels. Our pretrained StyleGAN is kept fixed during the caricature projection process. Much like the PSP network, we employ ``Mapper'', a small mapping network, which is trained to extract learned styles from the corresponding feature maps for each of the 14 target styles (for 256 x 256 images). This small mapping network is fully convolutional, downsampling the feature map to generate the corresponding 512-dimensional style input. It achieves this through a series of 2-strided convolutions followed by LeakyReLU activations. Specifically, the small feature map from the Mapper generates styles $W_{0} - W_{2}$, the medium feature map generates styles $W_{3} - W_{6}$, and the large feature map generates styles $W_{7} - W_{17}$.
We incorporate Restyle's \cite{alaluf2021restyle} iterative refinement method to enhance the reconstruction quality with each iterative step. We perform a single training iteration per batch with our model trained. The iterative outcome for $I_{real}$ is $I^{iter}_{real}$ and $I_{cari}$ is $I^{iter}_{cari}$.
To further transfer the background from $I_{real}$, a common approach involves applying a blending technique, as depicted in Equation \ref{eq: alpha}, as a post-processing step that swaps the inner face of $I^{f}_{real}$ and $I^{f}_{cari}$ with $I_{real}$. We execute the image matting technique, followed by the alpha blending process discussed in Section \ref{lb: Caricature dataset creation}, to produce our final projection $I^{B}_{real}$ and $I^{B}_{cari}$.

\begin{figure}[]
  \centering
  \includegraphics[width=\linewidth]{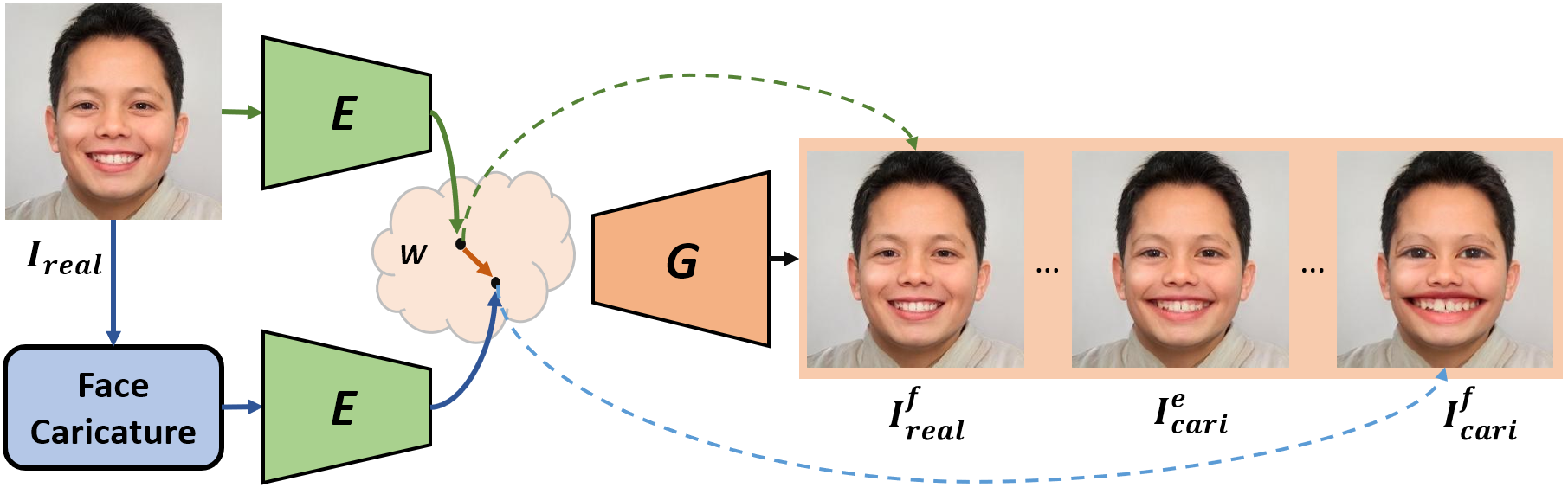}
  \caption{After training the encoder, we perform incremental caricature projection by performing latent walking from the real image toward the direction of the corresponding caricature image.}
  \label{fig: latent_projection}
\end{figure}

\subsubsection{Incremental Caricature Projection}

We utilize the disentangled nature of the StyleGAN latent space to create caricature faces of desired facial exaggeration. The disentangled latent spaces also facilitate smooth and predictable transitions between real and caricature faces. To perform the incremental facial exaggeration process, we employ our trained encoders, $E$, and our pretrained StyleGAN. The overview of our incremental facial exaggeration is shown in Figure \ref{fig: latent_projection}. 
Given an input image $I_{real}$, we create the corresponding caricature of the real face, $I_{cari}$. We fed the  $I_{real}$ and $I_{cari}$ to the encoder $E$. We Project two latent codes in the StyleGAN latent space $W$, one for $I_{real}$, represented as  $E(I_{real}) = z^{real}$ where $z^{real}$ is the real latent code, and another for $I_{cari}$, represented as $E(I_{cari}) = z^{cari}$ where $z^{cari}$ is the caricature latent code. We perform a latent walk from $z^{real}$ to $z^{cari}$ with the objective of $I^{f}_{cari}$ = $G(z^{real} + n_{cari})$ if $n_{cari}$ = 1, where $n_{cari}$ is the incremental latent steps of $z^{cari}$ direction. We can perform a uniform iteration represented as $I^{e}_{cari}$. We visualize more results in Section \ref{lb: ex  Incremental Projection}.

\subsubsection{Losses}

To achieve our objective, we employ a variety of losses during the training of our encoder. We incorporate the non-saturating GAN loss \cite{GAN_NIPS2014} along with $R_1$ regularization \cite{mescheder2018training} as the adversarial loss, proposed in \cite{Nitzan_distangle}. 

The purpose of regularization is to encourage the encoder to produce latent-style vectors that are closer to the average latent vector. The formulation of the regularization loss is as follows:

\begin{equation}
\mathcal{L}_{reg} = \parallel G(E(x))- \bar{w} \parallel _{2},
\label{eq: reg}
\end{equation}

\noindent where $\bar{w}$ represents the average style vector obtained from our pre-trained generator.

\vspace{1mm}

\noindent We employ the pixel-wise L2 loss,

\begin{equation}
\mathcal{L}_{2} = \parallel x-G(E(x)) \parallel _{2},
\label{eq: L2}
\end{equation}

\vspace{1mm}

\noindent In order to preserve the perceptual similarity, we use LPIPS \cite{zhang2018unreasonable} loss. The image is preserved better \cite{guan2020collaborative} as compared to the traditional approach \cite{johnson2016perceptual}.

\begin{equation}
\mathcal{L}_{LPIPS} = \parallel F(x)- F(G(E(x))) \parallel _{2},
\label{eq: Lpips}
\end{equation}

\vspace{1mm}

To generate a caricature face that retains similar facial characteristics, we employ identity loss. This involves integrating a dedicated recognition loss, which assesses the cosine similarity between the resulting image and its source.

\begin{equation}
\mathcal{L}_{ID}(x) =  1 - \left \langle  A(x),A(G(E(x)))  \right \rangle,
\label{eq: ID}
\end{equation}

\vspace{1mm}
\noindent where A represents the pretrained ArcFace \cite{deng2019arcface} network.

\noindent Collectively, our overall loss function is presented as

\begin{equation}
\mathcal{L}(x) = \lambda_{1}\mathcal{L}_{2}(x) + \lambda_{2}\mathcal{L}_{LPIPS}(x) + \lambda_{3}\mathcal{L}_{ID}(x) + \lambda_{4}\mathcal{L}_{reg}(x),
\label{eq: overall loss}
\end{equation}

\vspace{1mm}

\noindent where $\lambda_{1}$, $\lambda_{2}$, $\lambda_{3}$, and $\lambda_{4}$ are constants defining the loss weights.

\vspace{1mm}

\begin{figure}[]
  \centering
  \includegraphics[width=\linewidth]{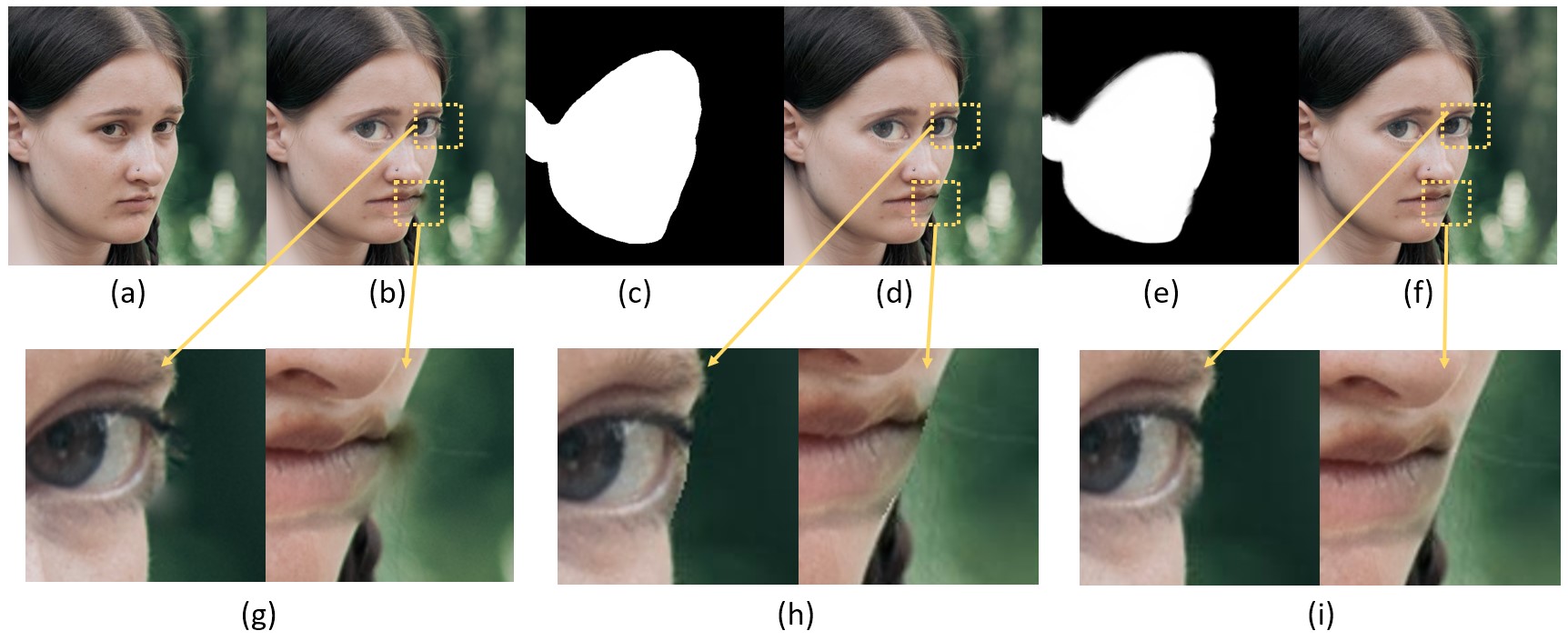}
  \caption{Experiment results with various masks during the blending process. (a) The real input face, (b) our face caricature dataset before blurring removal, (c) the facial mask obtained from image segmentation, (d) the blending of the original image with the facial mask, (e) the facial mask acquired through image matting, (f) the blending of the original image with the image matting mask, (g) blurring effect to the eyes and mouth region, (h) display of unnatural blending results, and (i) demonstrating the appropriate and natural blending of the image matting face mask with the original face.}
  \label{fig: blurring removal}
\end{figure}

\section{Implementation}
\label{lb: implementation}

\subsection{Dataset}

To showcase the efficacy of our approach, we produce caricature datasets and conducted experiments using a diverse dataset that encompasses two widely recognized datasets: FFHQ \cite{karras2019style} and CelebA-HQ \cite{liu2015deep}. 
The FFHQ dataset comprises 70,000 high-quality facial images, which we segmented into three groups based on the presence of eyeglasses: no glasses, reading glasses, and sunglasses. Specifically, we assigned approximately 56,500 images to the no-glasses group, 10,600 images to the reading glasses group, and 2,900 images to the sunglasses group.
Similarly, the CelebA-HQ dataset contains 30,000 high-quality facial images, and we also categorized these into three groups: no glasses, reading glasses, and sunglasses. Here, we allocated approximately 28,500 images to the no-glasses group, 1,000 images to the reading glasses group, and 500 images to the sunglasses group.

We use our new caricature and real faces from the FFHQ dataset for training our StyleGAN model. The FFHQ dataset's considerable size and high-quality image content render it suitable for effectively training a robust and representative caricature generator. 
For the encoder $E$ training, we use the FFHQ real and our new caricature dataset as the training set and the CelebA-HQ real and our new caricature dataset as the testing set.
By using these diverse datasets and splitting them into different groups based on eyeglass presence, we aimed to assess the ability of our approach to handle various scenarios and generate accurate caricatured faces across different styles and eyeglasses.

\subsection{Implementation Details}
We trained a StyleGAN model \cite{karras2019style, karras2020training} using real and our caricature datasets. The input and output image resolution for our caricature generation task was set to 256 x 256 pixels since our hardware resources are limited. The training process for the StyleGAN model was conducted on four Nvidia Titan Xp GPUs, each with 12 GB of RAM. It took approximately eight days to train the model using a batch size 16.
For the encoder $E$ training, we utilized the ResNet-IRSE50 architecture from Arcface \cite{deng2019arcface}, a pretrained model commonly used for facial recognition tasks.
In our $E$ training process, we set the values of the constants $\lambda$ as follows: $\lambda_{1}$ = 1, $\lambda_{2}$ = 0.8, $\lambda_{3}$ = 0.5, and $\lambda_{4}$ = 0.005. These constants were used to control and balance different aspects of the training process. We set other training details the same as \cite{richardson2021encoding}.

\begin{figure}[]
  \centering
  \includegraphics[width=\linewidth]{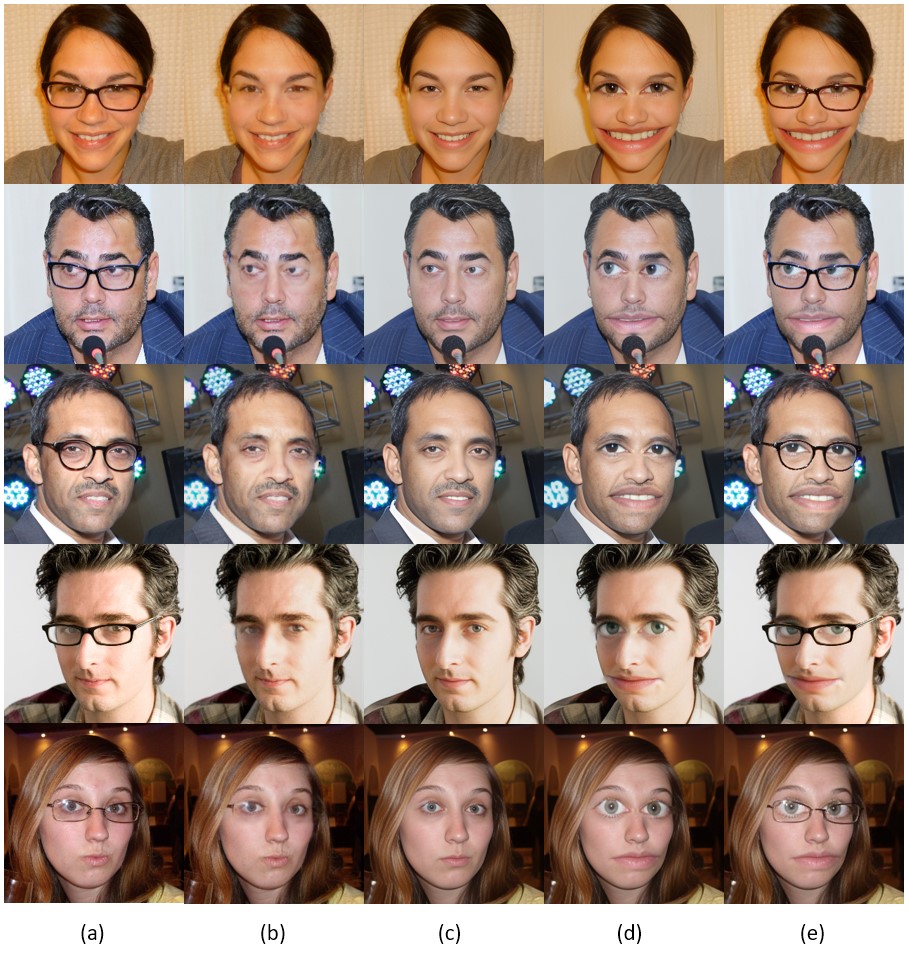}
  \caption{Results of various stages in the process of generating a reading glass caricature dataset: (a) The real input face, (b) the result after removing reading glasses and cast shadows, (c) the result after the face correction technique, (d) after applying our face caricature, and (e) the reading glass caricatured face.}
  \label{fig: glass removal step}
\end{figure}

\section{Experiments}
\label{lb: experiment}

\subsection{Experiments on Caricature generation}

The process of patch blending is of utmost importance in creating the face caricature dataset. The quality of the StyleGAN-generated images greatly depends on the seamless blending of these patches. However, when dealing with extreme head poses, the blending process can sometimes lead to blurriness.
To address this blurring issue, we employ a face mask that eliminates all blurriness, resulting in a more natural-looking image. Additionally, we introduce a face mask in conjunction with a matting mask, and we compare the outcomes, as demonstrated in Figure \ref{fig: blurring removal}. The image matting mask successfully eliminates blurriness along the facial contours, yielding a more natural appearance than just the face mask. It's worth noting that the face segmentation mask tends to produce unnatural edges, which can adversely impact the final result.

While removing reading glasses, the facial details concealed behind the glasses are inevitably lost. To address this issue, we employ a correction technique to recover the lost information. We illustrate the various stages of the reading glass removal process in Figure \ref{fig: glass removal step}. After the eyeglass removal, a significant portion of the information in the eye region is degraded, which can adversely impact caricature generation. However, the correction method not only restores both quality and fidelity but also exhibits remarkable resilience in handling the deteriorated portions, ultimately enhancing the creation of a superior caricature dataset.

\begin{figure}[t]
  \centering
  \includegraphics[scale=0.35]{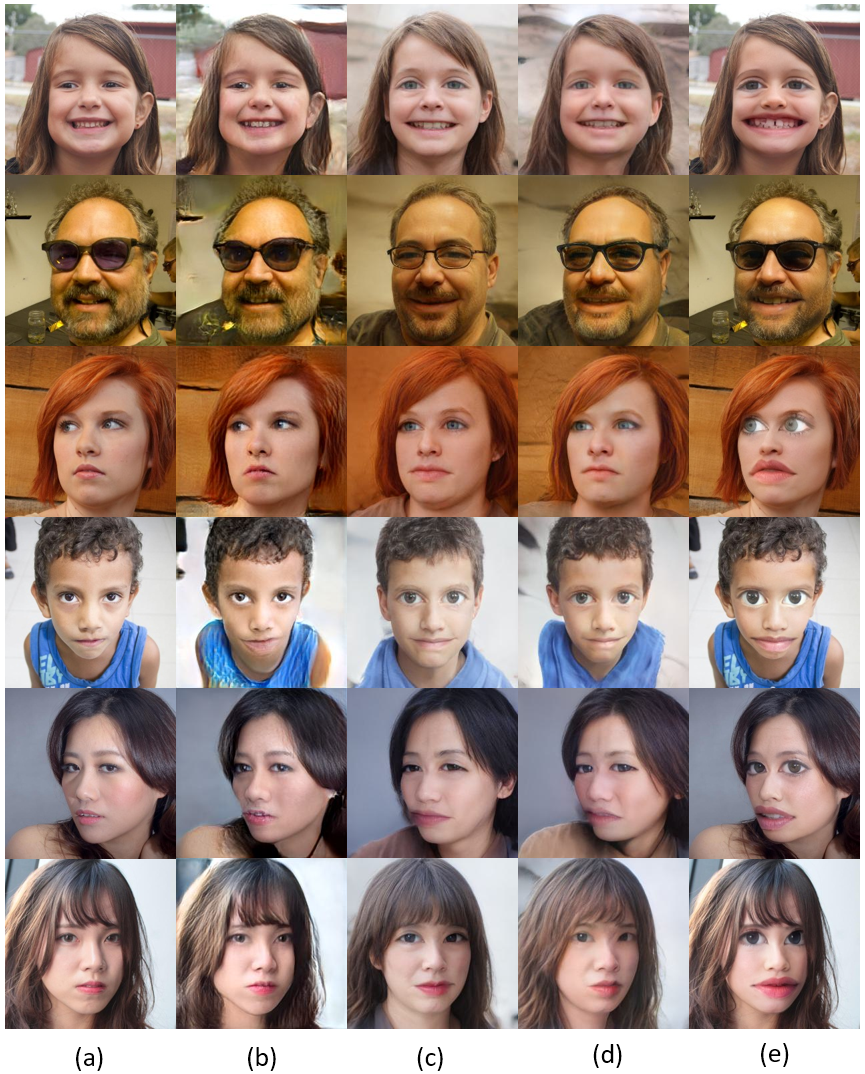}
  \caption{The projection result of the encoder trained with real and caricature images with background blending. (a) The real input face, (b) The real projected face, (c) The caricature input face, and (d) The caricature projected face.}
  \label{fig: real cari projection}
\end{figure}

\begin{table}[]
\caption{Comparison between the real and caricature projections results.}
\label{table: real cari projection}
\begin{tabularx}{\columnwidth}{@{} l *{10}{C} c @{}}
\toprule
Projection                    & $\downarrow$ LPIPS & $\downarrow$ L2    & $\uparrow$ SSIM \\
\midrule
Real Face Projection        & 0.049 & 0.008 & 0.91 \\
Caricature  Face Projection & 0.056 & 0.009 & 0.90  \\
\bottomrule
\end{tabularx}
\end{table}

\subsection{Experiments on Caricature Projection}

Following the training of our caricature projection using the encoders trained on real and our caricature datasets, we conducted a series of experiments to assess the efficacy of our projection method. We evaluate the encoder results by comparing the input and output faces for the real and caricature images, as illustrated in Figure \ref{fig: real cari projection}. The outcome of projecting real and caricature images demonstrates the effectiveness of our approach, as the input and the resulting projected images exhibit minimal differences. It is evident that our real and caricature projections consistently yield attractive and aesthetically pleasing outcomes. Table \ref{table: real cari projection} shows the evaluation of our projection results.

\begin{figure}[]
  \centering
  \includegraphics[scale=0.305]{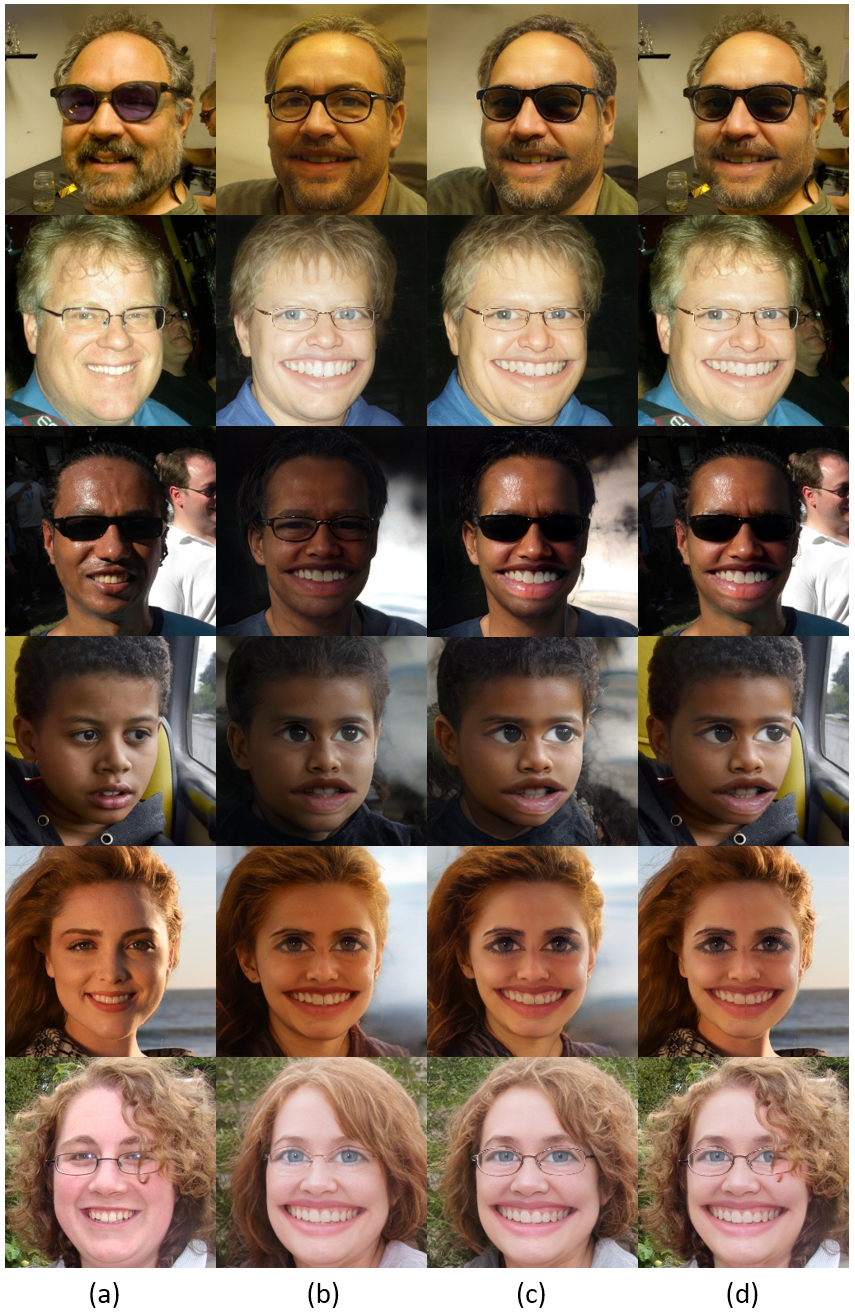}
  \caption{Result of our iterative steps during our caricature projection. (a) The real input face, (b) the initial projected face after caricature creation from the corresponding real face, (c) the final iterative step, and (d) our projected caricature face (with background blending).}
  \label{fig: iterative step}
\end{figure}

%comparison of iterative steps
During our projection method, each iterative step enhances the image quality, as illustrated in Figure \ref{fig: iterative step}. The iterative process enhanced the eyeglasses information, as demonstrated in rows 1, 2, 3, and 6. It's noteworthy that there is a substantial improvement in head pose and facial expression, as observed in row 4. There is also an improvement in the lighting and skin color resemblance with the input image, as observed in row 5.

\subsection{Experiments on Incremental Projection}
\label{lb: ex  Incremental Projection}

\begin{figure}[]
  \centering
  \includegraphics[width=\linewidth]{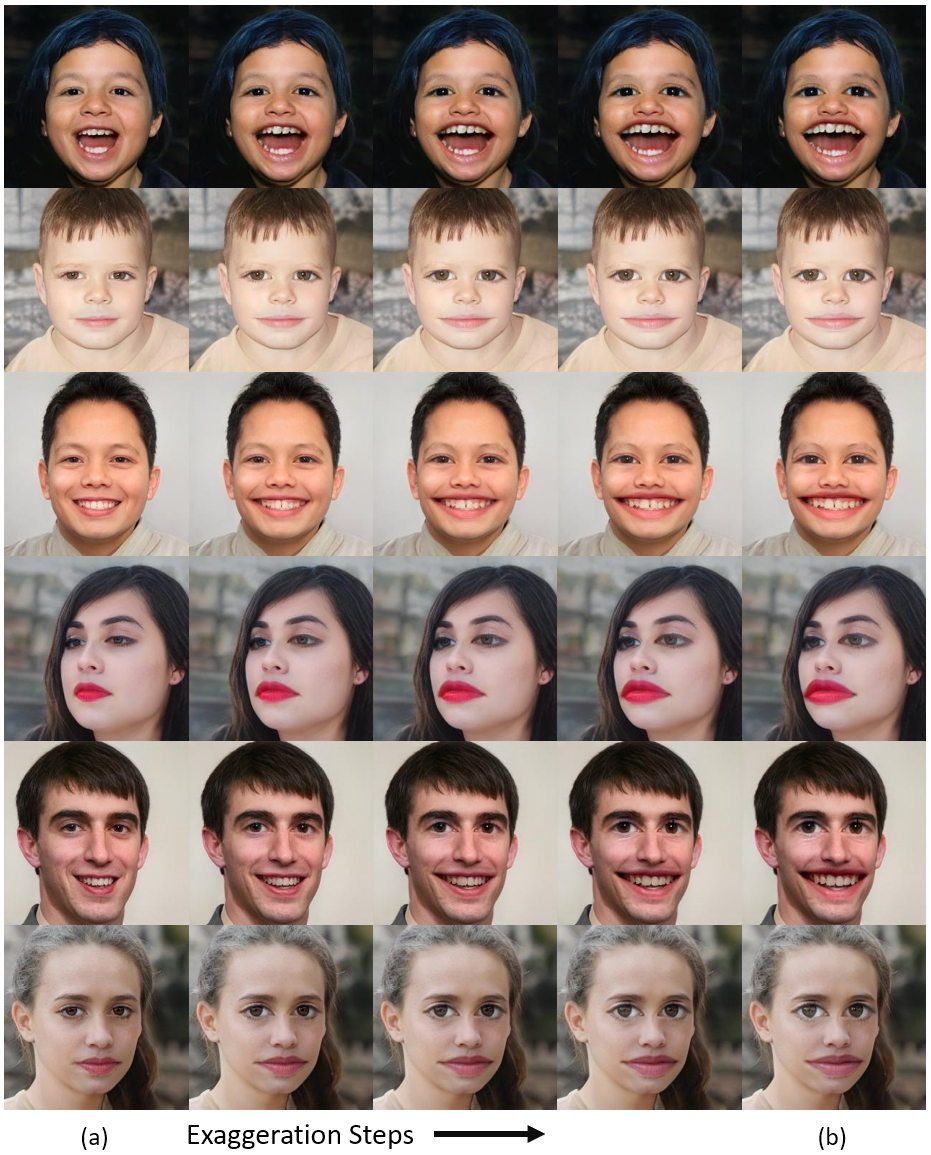}
  \caption{Result of our incremental facial exaggeration steps from the real face to the corresponding caricature face. (a) The real input face, and (b) the final caricature face.}
  \label{fig: Incremental Projection}
\end{figure}

We perform an incremental caricature projection method in which we gradually exaggerate facial features, as demonstrated in Figure \ref{fig: Incremental Projection}. The visual result shows that the exaggeration affects the eyes and mouth, leaving all other facial attributes unchanged. The exaggeration steps are crucial in our method as they hold great significance in our approach, as the extent to which individuals prefer facial exaggeration varies. This step provides flexibility and robustness in our caricature projection process.

\begin{figure}[]
  \centering
  \includegraphics[width=\linewidth]{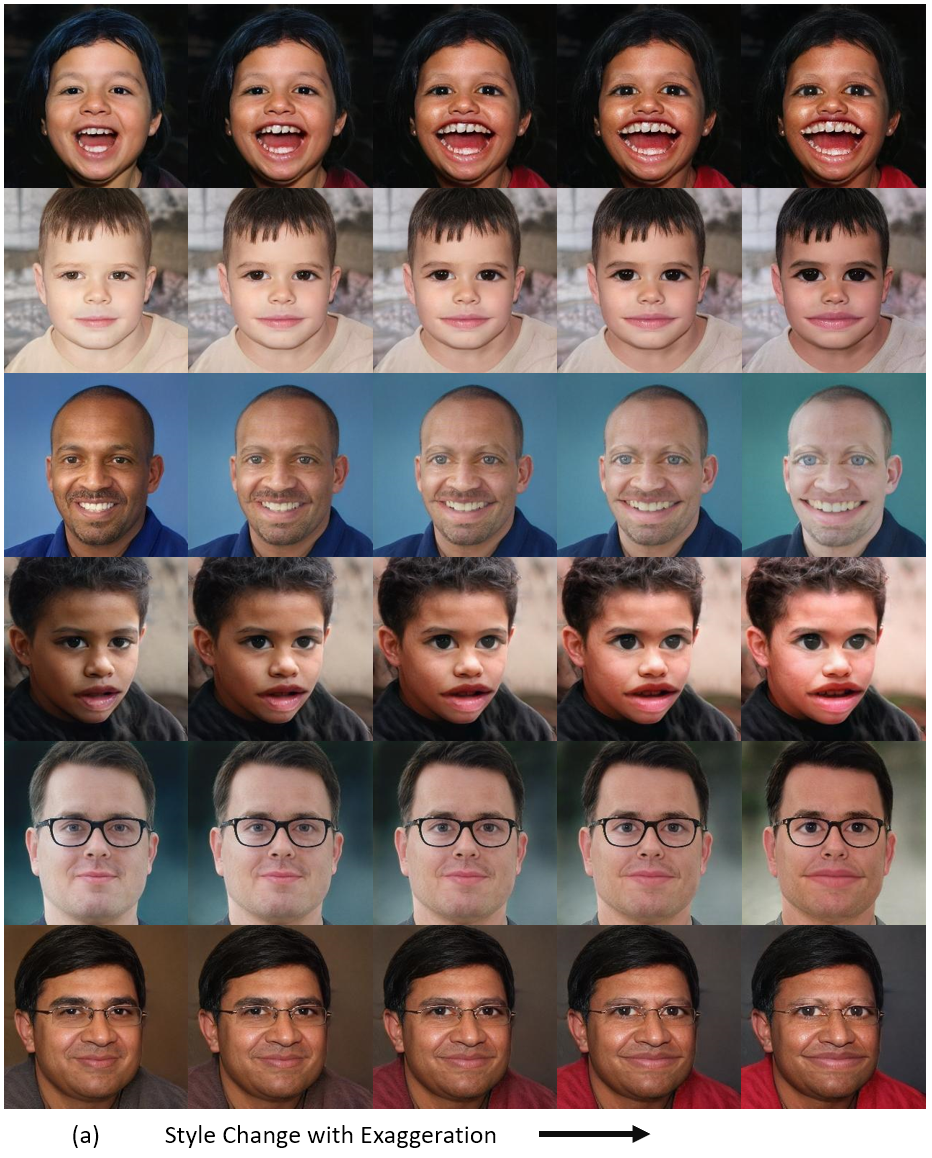}
  \caption{Result of our incremental facial exaggeration steps with desirable style change from the real face to the corresponding caricature face. (a) The real input face.}
  \label{fig: style_mixing}
\end{figure}

Furthermore, we introduce a style-mixing element into the exaggeration process. During the exaggeration process, we can select and incorporate the desired style, as shown in Figure \ref{fig: style_mixing}. The desired style can be blended by mixing the style codes within the finer layers of StyleGAN.

\subsection{Comparison to state-of-the-art method}
%comparison of caricature projection 

\begin{figure}[t]
  \centering
  \includegraphics[width=\linewidth]{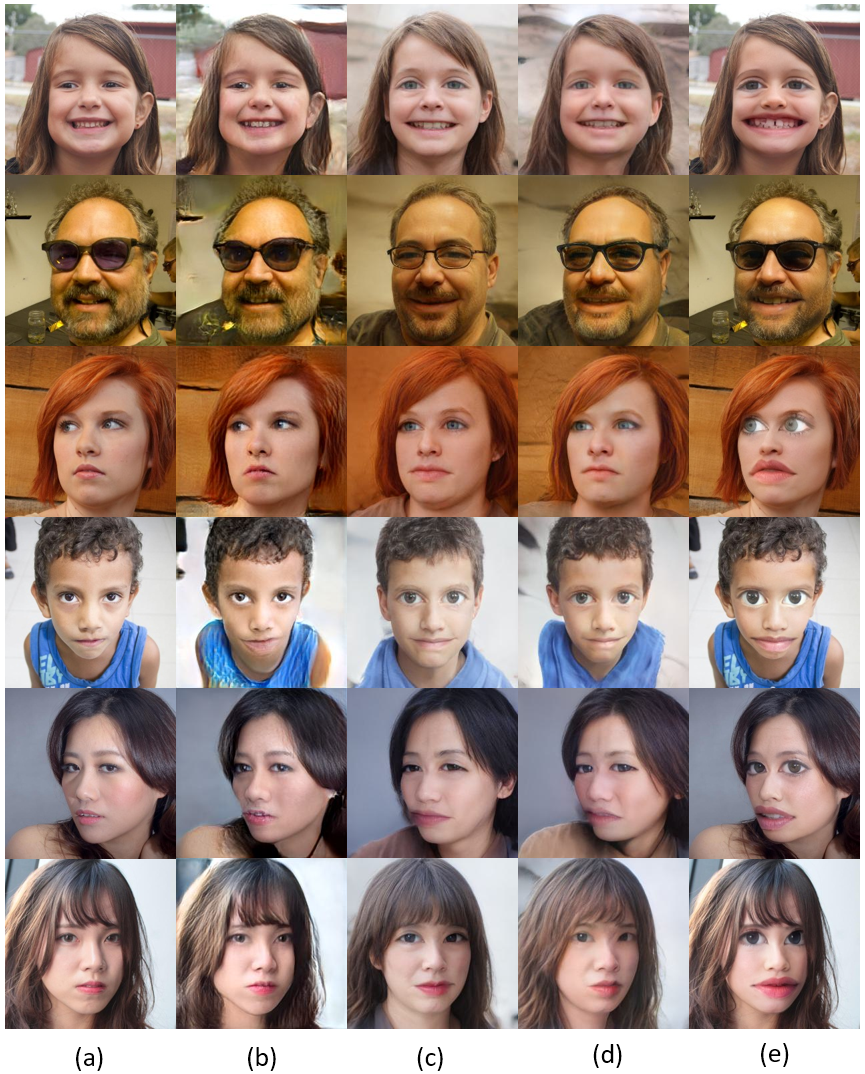}
  \caption{Qualitative comparison results of different face projection techniques using the StyleGAN which is trained on our caricature dataset. (a) The real input face, (b) hyperstyle encoder, (c) e4e encoder, (d) restyle encoder and (e) our projected caricature face.}
  \label{fig: compare_projection}
\end{figure}

We evaluate the performance of our caricature projection method by comparing it to the state-of-the-art technique. This comparison encompasses all the encoder-based to assess the efficacy of our method comprehensively. In our qualitative evaluation, as depicted in Figure \ref{fig: compare_projection}, we conduct experiments utilizing various approaches. We explore three encoder-based methods: Hyperstyle \cite{alaluf2022hyperstyle}, e4e \cite{tov2021designing} and Restyle \cite{alaluf2021restyle}. These encoders are trained using a pretrained StyleGAN, which is trained using only our caricature images. 
Hyperstyle tends to generate caricatures that closely resemble the original image in terms of structure, as it tunes the StyleGAN weights to retrieve the original image rather than caricature faces. 
Conversely, the caricatures produced by the e4e encoder yield superior caricature results when compared to Hyperstyle, and the results provide convincing caricature results. The Restyle results resemble more the real images than the caricature faces. 
Finally, our method outperforms all techniques, particularly the facial exaggeration and the expression and head pose, which more closely resemble those of the real image. Furthermore, our approach excels at handling occluded faces and produces caricatures that more closely resemble the original eyeglasses.
Overall, our approach demonstrates superior results compared to existing techniques, offering a more faithful representation of the original image's characteristics while achieving high-quality caricature results.

\begin{figure}[t]
  \centering
  \includegraphics[width=\linewidth]{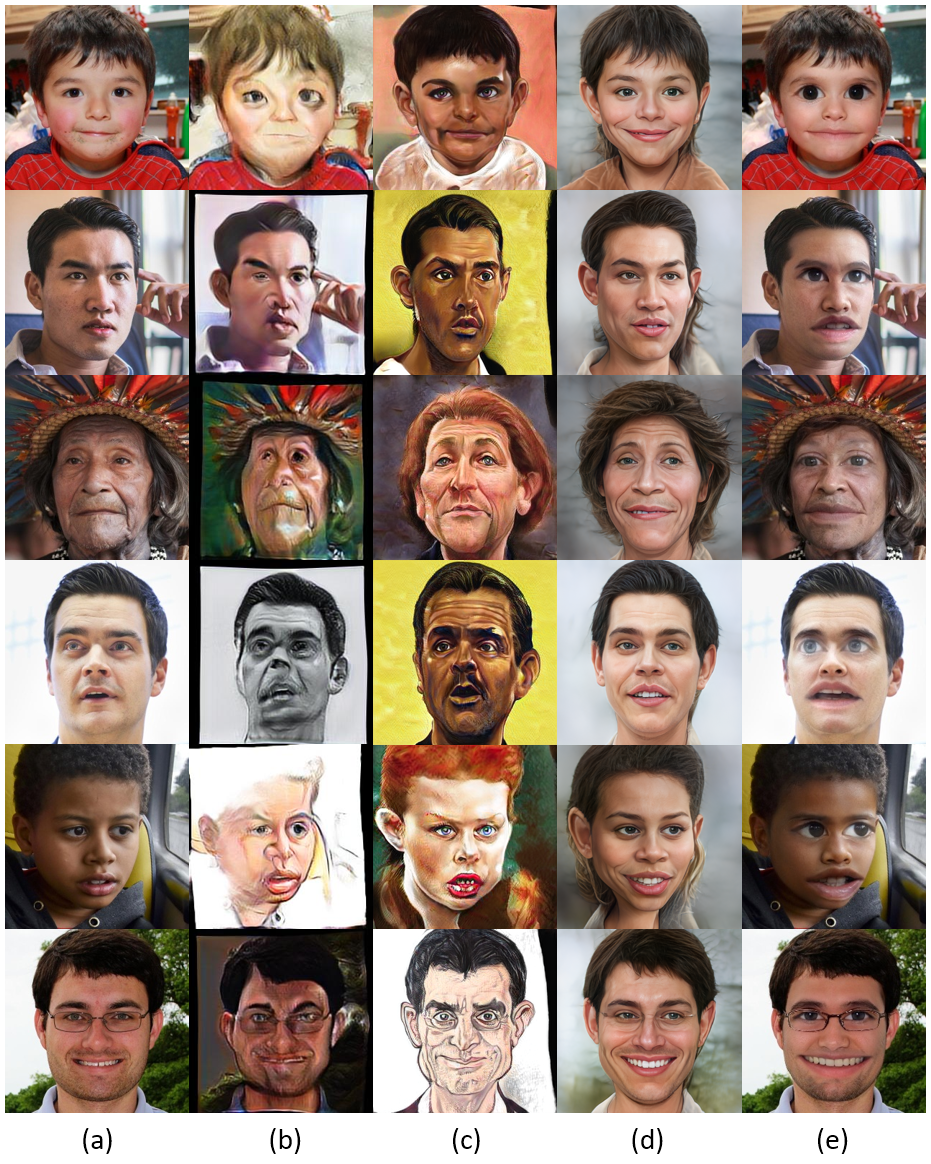}
\caption{Qualitative comparison results of different face caricature methods. (a) The input real face, (b) WarpGAN, (c) StyleCariGAN, (d) DualStyleGAN, and (e) our projected caricature face.}
  \label{fig: compare_caricature}
\end{figure}

Moreover, we conducted a qualitative analysis of various state-of-the-art caricature methods, comparing our outcomes with those of WarpGAN \cite{shi2019warpgan}, StyleCariGAN \cite{jang2021stylecarigan}, and DualStyleGAN \cite{yang2022pastiche}, as shown in Figure \ref{fig: compare_caricature}. All results were generated using the pretrained models provided by the respective authors using CelebA-HQ \cite{liu2015deep}.
The WarpGAN struggled to produce caricatures with proper facial structures and weakly stylized images. The StyleCariGAN had difficulty preserving the original image's identity and heavily relied on the chosen style. The DualStyleGAN yielded convincing results but was limited in retaining the original attributes.
In contrast, our caricature results excelled in quality, maintaining both style and the facial attributes of the original image. The exaggeration achieved in our projected caricature faces holds promise for practical applications in real-world scenarios. 
We also performed a quantitative evaluation to assess the degree of resemblance between real and caricature images, as shown in Table \ref{table: compare}. The identity similarly calculation uses the ArcFace \cite{deng2019arcface} method. The Result shows that our method demonstrated the most favorable score, making it effectively exaggerate facial features and align with our primary goal of making it applicable in real-world scenarios.

\begin{figure}[]
  \centering
  \includegraphics[width=\linewidth]{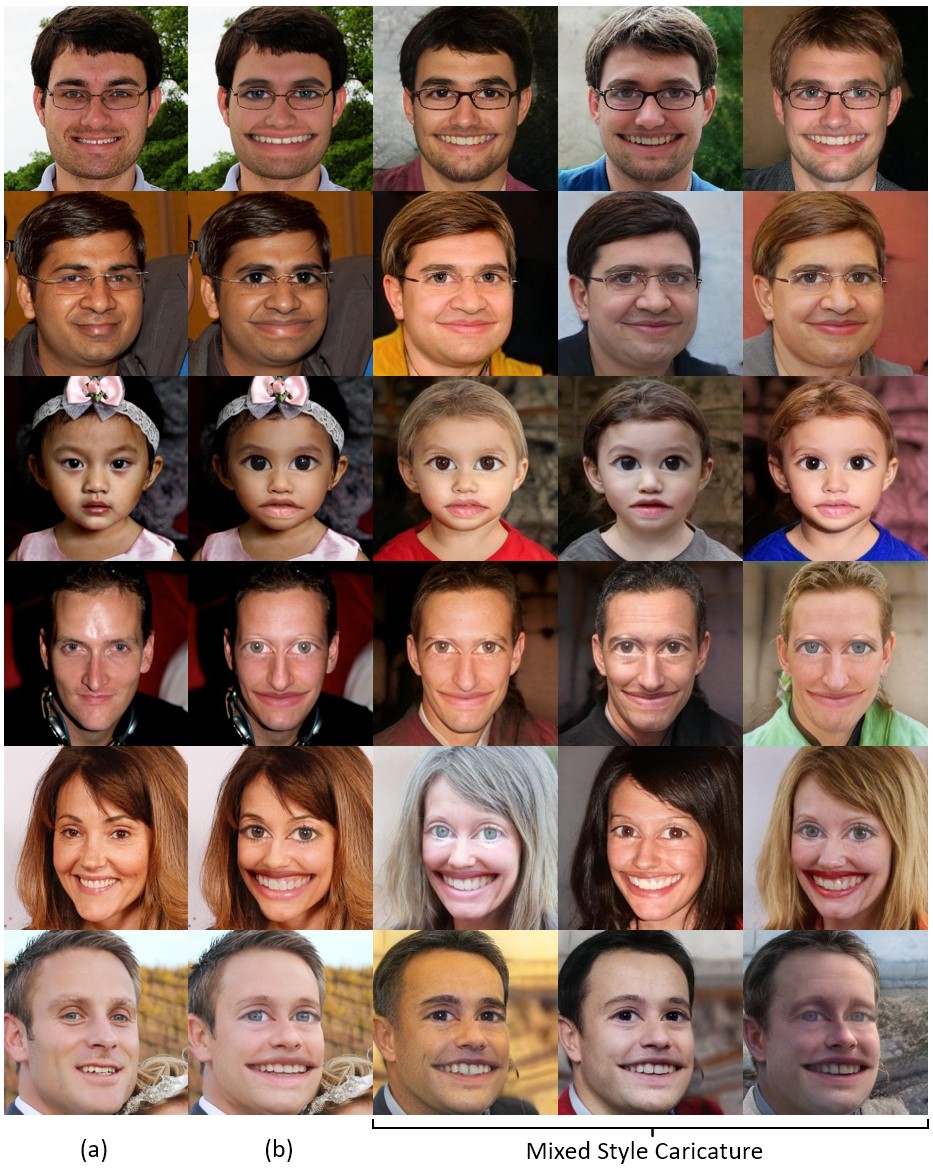}
  \caption{Visual results obtained using our method. (a) The input real face, (b) our projected caricature face, and mixed style caricature faces.}
  \label{fig: final}
\end{figure}

%display of real faces and generated faces
We showcase the outcomes of our approach using various facial images captured in diverse conditions, as shown in Figure \ref{fig: final}. Our methodology consistently delivers outstanding facial caricature results marked by realism and the retention of the original facial attributes. We visualize the generation of different style types that can be incorporated with our caricature face. Moreover, our method generates faces with occlusions, such as reading glasses and sunglasses. Furthermore, it displays versatility by producing caricatured faces across different age groups and adapting to various artistic styles.

\begin{table}[]
\caption{Quantitative results for different caricature creation methods.}
\label{table: compare}
\begin{tabularx}{\columnwidth}{@{} l *{10}{C} c @{}}
\toprule

Method & $\uparrow$Identity & $\downarrow$FID & $\downarrow$LPIPS & $\downarrow$L2  & $\uparrow$SSIM   \\
\midrule
        
WarpGAN \cite{shi2019warpgan}   & 0.26 & 74.60   & 0.48     & 0.65  & 0.25    \\
StyleCariGAN \cite{jang2021stylecarigan}  & 0.11     & 52.35   & 0.47     & 0.41 & 0.32      \\
DualStyleGAN \cite{yang2022pastiche}            & 0.08    & 104.51    & 0.42    & 0.51   & 0.33      \\
Our      & \textbf{0.37} & \textbf{38.08}      & \textbf{0.06}     & \textbf{0.01}    & \textbf{0.85}  \\

\bottomrule
\end{tabularx}
\end{table}

\section{Conclusion}
\label{lb: conclusion}

In this paper, We generate realistic facial caricatures featuring exaggerated features suitable for real-world applications. Our methodology is carefully crafted to emphasize exaggerating the eyes and mouth while preserving the original facial contours. 
We have introduced an innovative caricature generation method that comprises two key stages: face caricature generation and face caricature projection. In the face caricature generation phase, we construct caricature datasets using real images. Subsequently, we train a styleGAN to synthesize various styles of real and caricatured faces.
The face caricature projection step takes input images and transforms them into corresponding real and caricatured faces. Our caricature projection process excels at producing highly realistic results while faithfully retaining the original facial attributes and identity. We also perform an incremental caricature projection method in which we gradually exaggerate facial features. We emphasize the importance of the exaggeration steps in our technique because different people have varying preferences regarding facial exaggeration. This process gives our caricature projection process flexibility and resilience. 
Our caricatures stand out in their superior realism and quality compared to previous methods. They offer visually convincing results suitable for real-world applications.

\begin{figure}[h]
  \centering
  \includegraphics[width=\linewidth]{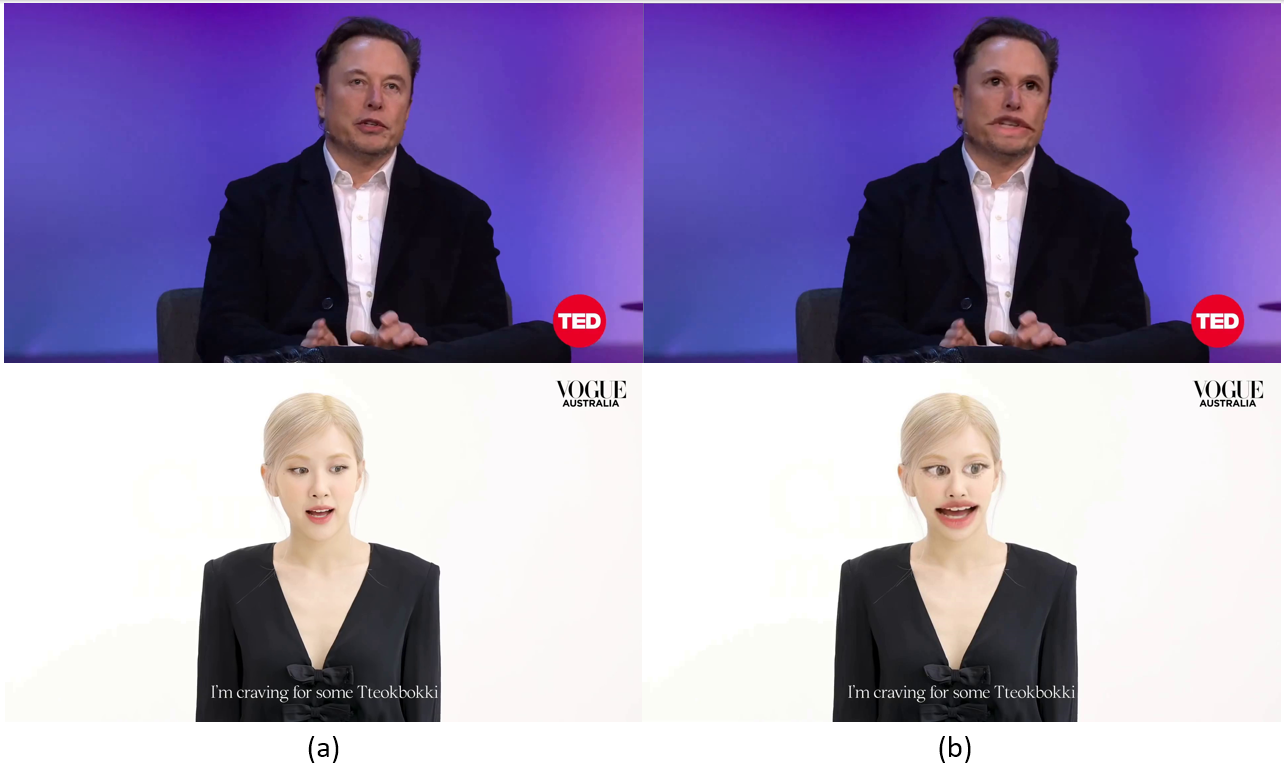}
  \caption{Visualization for protecting the privacy of an individual using our caricature faces in a full image. (a) The input real face, and (b) our projected caricature face.}
  \label{fig: projection_image}
\end{figure}

Our approach introduces an innovative method for crafting exaggerated facial representations while maintaining a realistic style. Our future work includes exploring face de-identification using our caricature-projected faces to conceal the important facial features that can be used to identify the individual. Our core concept revolves around using caricature faces to protect individuals' privacy. We illustrate one example of protecting the privacy of an individual using our caricature faces in Figure \ref{fig: projection_image}.

\subsection{Limitations}
It's crucial to explore the adaptability of our method in various applications. However, it's essential to acknowledge that our approach is tailored to specific applications and does have inherent limitations when applied in broader contexts. Notably, we emphasize exaggerating features in the eyes and mouth region, limiting the range of caricatures we can generate. Additionally, our method relies on real images, restricting its stylistic diversity and making it less suitable for producing different out-of-domain caricatures.
Our method is entirely automated, and future improvements could involve enhancing controllability through additional caricature examples or user interaction.
Nonetheless, our approach holds great promise for specific applications, and we are eager to refine and expand its capabilities in the future.

% \appendices

% % you can choose not to have a title for an appendix
% % if you want by leaving the argument blank
% \section{}

\section*{Acknowledgment}

This work was supported by the Institute of Information \& communications Technology Planning \& Evaluation (IITP) grant funded by the Korean government (MSIT) (No. 2019-0-00203, Development of 5G-based Predictive Visual Security Technology for Preemptive Threat Response) and also by the MSIT (Ministry of Science and ICT), Korea, under the Innovative Human Resource Development for Local Intellectualization support program (IITP-2022-RS-2022-00156389) supervised by the IITP (Institute for Information \& communications Technology Planning \& Evaluation).

\ifCLASSOPTIONcaptionsoff
  \newpage
\fi



%\bibliographystyle{IEEEtran}
\bibliographystyle{IEEEtran}
\bibliography{mybib}


% \end{IEEEbiography}


% % insert where needed to balance the two columns on the last page with
% %\newpage



%\vfill


\end{document}